\documentclass[10pt,twocolumn,letterpaper]{article}

\usepackage{cvpr}              % To produce the CAMERA-READY version
\usepackage{xcolor}         % colors
\usepackage{subcaption}

\usepackage[export]{adjustbox}
\usepackage{graphicx}
\usepackage{enumitem}
\usepackage{float} 
\usepackage{placeins}

\usepackage{booktabs}
\usepackage{utfsym}
\newcommand{\crossmark}{\scalebox{0.75}{\usym{2613}}}
\usepackage{wrapfig}
\usepackage{multirow}
\usepackage{amsmath}
\usepackage{caption,setspace}
\usepackage{nicefrac}
\usepackage{amssymb}
\usepackage{bm}

\renewcommand{\paragraph}[1]{\noindent\textbf{#1}}
\newcommand{\fref}[1]{Figure~\ref{#1}}
\newcommand{\sref}[1]{Section~\ref{#1}}
\newcommand{\tref}[1]{Table~\ref{#1}}

\definecolor{cvprblue}{rgb}{0.21,0.49,0.74}
\usepackage[pagebackref,breaklinks,colorlinks,allcolors=cvprblue]{hyperref}

\def\paperID{14350} % *** Enter the Paper ID here
\def\confName{CVPR}
\def\confYear{2025}

\title{SuperPC: A Single Diffusion Model for Point Cloud \\ Completion, Upsampling, Denoising, and Colorization}

\author{
Yi Du, \space
Zhipeng Zhao, \space
Shaoshu Su, \space
Sharath Golluri, \space
Haoze Zheng, \space
Runmao Yao, \space
Chen Wang \\
Spatial AI \& Robotics (SAIR) Lab, University at Buffalo \\
{\tt\small \{yid, chenw\}@sairlab.org}
}

\begin{document}
\maketitle

\begin{abstract}
Point cloud (PC) processing tasks—such as completion, upsampling, denoising, and colorization—are crucial in applications like autonomous driving and 3D reconstruction. 
Despite substantial advancements, prior approaches often address each of these tasks independently, with separate models focused on individual issues. 
However, this isolated approach fails to account for the fact that defects like incompleteness, low resolution, noise, and lack of color frequently coexist, with each defect influencing and correlating with the others.
Simply applying these models sequentially can lead to error accumulation from each model, along with increased computational costs. 
To address these challenges, we introduce SuperPC, the first unified diffusion model capable of concurrently handling all four tasks. Our approach employs a three-level-conditioned diffusion framework, enhanced by a novel spatial-mix-fusion strategy, to leverage the correlations among these four defects for simultaneous, efficient processing.
We show that SuperPC outperforms the state-of-the-art specialized models as well as their combination on all four individual tasks.
Project website: \href{https://sairlab.org/superpc/}{\textcolor[rgb]{0.3,0.5,1}{https://sairlab.org/superpc/}}.
\end{abstract}    
\vspace{-10pt}
\section{Introduction}
\label{sec:intro}
\vspace{-3pt}
\begin{figure*}[ht]
    \centering
    \includegraphics[width=1\linewidth]{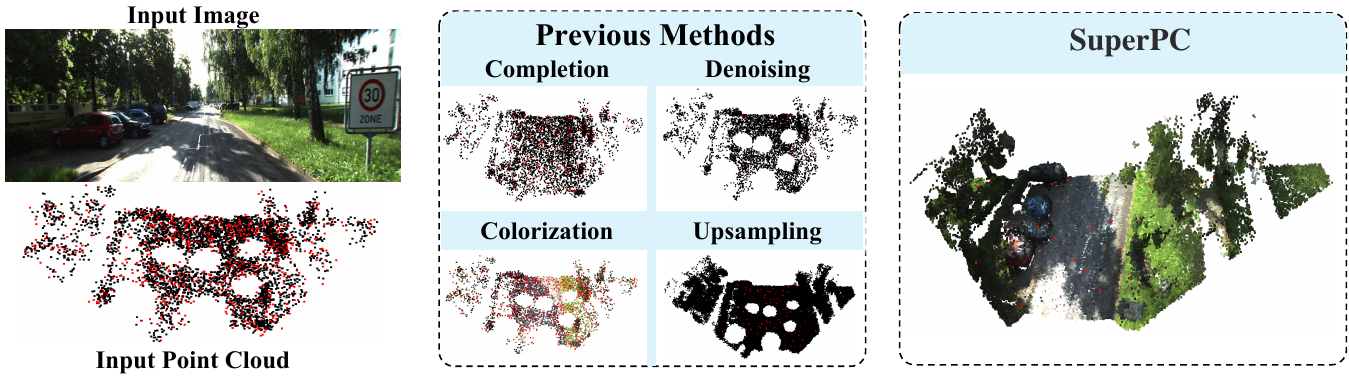}
    \vspace{-18pt}
    \caption{We propose SuperPC, a novel neural architecture that jointly solves inherent shortcomings in the raw point clouds, including noise, sparsity, incompleteness, and the absence of color. To the best of our knowledge, it is the \textbf{first} single diffusion model that can simultaneously tackle the \textbf{four} major challenges in the field of point cloud processing. Red points denote high noise for visualization.}
    \label{fig:SuperPC_intro2}
    \vspace{-3pt}
\end{figure*}

Point cloud (PC) processing \cite{leberl2010point,guo2020deep} has become a cornerstone technique in many fields such as object recognition \cite{aldoma2012tutorial,zhou2018voxelnet}, mapping \cite{schwarz2010mapping,zhang2014loam}, autonomous vehicle navigation \cite{li2020lidar,ort2018autonomous}, and 3D modeling \cite{remondino2003point,diaz20153d}. This diverse applicability stems from the ability of point cloud to capture the geometric intricacies of the physical world with high fidelity, offering a detailed substrate from which nuanced computational analyses can be performed \cite{leberl2010point,rusu20113d,bisheng2017progress,guo2020deep}.

Despite this critical need and importance, the prior PC processing methods, either geometry-based or data-driven, have typically concentrated on individual sub-tasks like denoising \cite{roveri2018pointpronets, luo2021score}, upsampling \cite{yu2018pu, he2023grad}, completion \cite{yuan2018pcn, yu2023adapointr}, and colorization \cite{shinohara2021point2color, liu2019pccn}, rather than a single comprehensive solution.
While some studies \cite{lyu2021conditional, choe2021deep, melas2023pc2, luo2021score} addressed the partial combination of those tasks, there is still a lack of a single model that addresses all four tasks simultaneously, which could bring multiple benefits to this field.
This is because an integrated model may be advantageous not just for its computational efficiency but also for preventing error accumulation and allowing them to implicitly improve each other due to the interconnectivity of the tasks.
For instance, in the experiments, we found that the error of the completion model accumulated and affected the upsampling; the error pattern produced by an upsampling model often resulted in suboptimal denoising.
As a result, we argue that a single unified model that is capable of concurrently handling the four point cloud processing tasks is necessary.

To this end, we take PC processing as generative tasks and resort to diffusion-based models \cite{saharia2022image, lugmayr2022repaint, luo2021diffusion, melas2023pc2, lyu2021conditional, lugmayr2022repaint}, which have shown significant advancements on PC generation.
Despite this, using diffusion models to address all four tasks is challenging, and previous studies \cite{luo2021diffusion, melas2023pc2, lyu2021conditional, lugmayr2022repaint} have been limited to tackling either individual tasks. 
% In this work, we find one reason is that previous methods only preserve part information (global, local, or raw) of the input data as a condition to guide the diffusion model, potentially leading to a biased diffusion process.
% A potential reason is that prior models preserve only part of the input information, whether it is global, local, or raw, taking input PC as a condition to guide the diffusion process.
The first limitation is their selective preservation of input information, whether global, local, or raw, using the input point cloud solely as a condition to guide the diffusion process. These approaches lacks the ability to generate point clouds that meet the requirements of all four tasks, as each task possesses distinct characteristics and demands.
For instance, the approach \cite{luo2021diffusion} relies solely on the global shape latent features for guiding 3D shape generation, which results in a limited capacity to capture the local object details; ``$\text{PC}^2$'' \cite{melas2023pc2} only employs the raw features of the input image as the condition, lacking the ability to preserve global-level information.
Another limitation of previous methods is their restriction to a single modality. For instance, the approach by Luo et al. \cite{luo2021diffusion} is confined to processing point cloud inputs, while the method by Luke et al. \cite{melas2023pc2} is limited to handling only image inputs.
As a result, they only work for simple object-level PC like ShapeNet \cite{chang2015shapenet} but struggle on complex scene-level PC such as KITTI-360 \cite{liao2022kitti}.

To address the first limitation, we introduce SuperPC, which integrates a Three-Level-Conditioned (TLC) diffusion framework to simultaneously handle four tasks. 
% We also employ a novel Spatial-Mix-Fusion (SMF) strategy—loose coupling early fusion and deep fusion \cite{huang2022multi}—to integrate images and point clouds, fully leveraging the fused information as three-level conditions to guide PC generation.
For tackling the second limitation, we propose a novel Spatial-Mix-Fusion (SMF) strategy to integrate images and point clouds.
It loosely couples the early-fusion strategy and the deep-fusion strategy \cite{huang2022multi}, fully leveraging the fused information as three-level conditions to guide PC generation.
Additionally, we design three modules, namely, a raw, a local, and a global module as the three conditions to control the diffusion process.
Specifically, (1) the raw module preserves the original information of the input point cloud and image via projection and interpolation, which can also be viewed as an early-fusion method to combine the two input modalities; 
(2) the local module extracts geometrical and texture details by fusing the local feature maps of the two modalities, which can also be viewed as a deep-fusion method; 
and (3) the global module guarantees high-level semantic consistency by condensing the fused local feature maps into a latent code. 
This enables SuperPC to generate high-quality, color-rich point clouds across diverse tasks and datasets while faithfully preserving local texture and geometrical information.
To the best of our knowledge, SuperPC is the \textbf{first} single diffusion model fusing \textbf{two} modalities by utilizing \textbf{three}-level conditions to concurrently tackle the \textbf{four} PC processing tasks.
Our contributions can be summarized as follows:
\begin{itemize}[noitemsep,topsep=0pt,parsep=0pt,partopsep=0pt]
  \item We propose SuperPC, a three-level-conditioned diffusion model framework with a novel spatial-mix-fusion strategy, allowing a single model, for the first time, to effectively tackle all four major challenges in PC processing.
  \item Our SuperPC model not only demonstrates superior performance on combination task but also surpasses existing SOTA models across all four individual PC processing tasks on scene-level benchmarks while exhibiting impressive efficiency and generalization capability.
  % \item We conducted an experiment that, for the \textbf{first} time, demonstrates how a single, unified model outperforms any combination of multiple models for different point cloud processing tasks, underscoring the essential need for a unified approach to address this gap.
  \item We have designed three new benchmarks for PC processing tasks: one object-level and two scene-level benchmarks. Additionally, these benchmarks are organized into two evaluation tracks to assess generalization ability—Object-to-Scene and Sim-to-Real.
\end{itemize}

% \vspace{-8pt}
\section{Related Works}
\subsection{PC processing Tasks} \label{sec:four_tasks}
\vspace{-3pt}
\noindent\textbf{Point Cloud Upsampling}, essential for enriching sparse 3D scanning outputs, has evolved from traditional methods like Alexa's work \cite{alexa2003computing} to those deep learning-driven methods \cite{yu2018pu, yifan2019patch, li2019pu, feng2022neural, he2023grad, qu2023conditional, qu2024conditional}. However, these methods struggle to upsample and preserve the details of scene-level point clouds. Unlike previous efforts, SuperPC addresses this challenge by preserving geometric details through the three-level-condition framework. 
% Additionally, our method is capable of producing dense point clouds at any desired upsampling rate, a feat not achievable by most of prior studies.

\paragraph{Point Cloud Completion} is aiming to generate points for both unobserved areas and incomplete shapes, as initially proposed by pioneers in the field \cite{sung2015data, nguyen2016field}. Following research has employed deep learning techniques, resulting in substantial improvements in the quality of completed point clouds \cite{han20193d2seqviews, huang2020pf, alliegro2021denoise, zhang2021unsupervised}. Yu's methods \cite{yu2021pointr, yu2023adapointr} exemplify this progress by converting the point cloud to a sequence of point proxies. However, they struggle in overly sparse conditions and unseen scenarios since they cannot extract dense structural and semantic information. In contrast, SuperPC can effectively fuse and utilize information from both the input point clouds and images, which contain dense semantic features via the novel spatial-mix-fusion strategy.

\paragraph{Point Cloud Denoising} is critical for refining the point cloud by removing outliers. Previous efforts \cite{lee2000curve, roveri2018pointpronets, rakotosaona2020pointcleannet, luo2021score} try to tackle it as a single task with traditional filter-based and novel learning-based methods. However, this process is heavily dependent on the point cloud upsampling and the completion tasks. For instance, denoising models may mistakenly remove a table leg because they recognize it for outliers when the table point cloud is too sparse or incomplete.

\paragraph{Point Cloud Colorization}:  
% Having achieved a dense, complete, and clean point cloud, the subsequent challenge is colorizing these points accurately. 
Traditional projection methods have been supplemented by geometry-based studies \cite{chan2021post,cho2014efficient,crombez20153d}, yet they struggle to color unseen 
and occluded points. 
Deep learning advancements \cite{liu2019pccn,hou2019range,gao2023scene,shinohara2021point2color} have enabled colorization solely based on geometry without image input, but often at the cost of generating colors that diverge from reality. Specialized models like Takayuki's work \cite{shinohara2021point2color} focus on niche applications such as Airborne LiDAR point clouds colorization and cannot be applied to everyday objects and scenes.
% In contrast, SuperPC incorporates the three-level-condition framework to fuse the two modalities. It can render both visible and unseen areas of the point cloud with realistic colors based on real-world images.
In contrast, SuperPC can render both visible and occluded point clouds with realistic colors based on the input image and the point cloud geometric features for both object-level and scene-level point clouds on both simulation and real-world datasets.

\vspace{-5pt}
\subsection{Diffusion Models for Point Cloud}
\vspace{-5pt}
The diffusion model, as evidenced by several studies \cite{saharia2022image, lugmayr2022repaint, ramesh2021zero}, has shown significant capabilities in image synthesis, prompting its application in 3D point cloud challenges. 
Luo \cite{luo2021diffusion} pioneered the use of a conditional diffusion model for 3D PC generation, guided by the shape latent condition of the input cloud. 
Building on this, Zhuo \cite{zhou20213d} and Zeng \cite{zeng2022lion} explored unconditional shape generation and the hierarchical point diffusion model for point cloud synthesis in latent space. 
Besides the point cloud synthesis, some researchers \cite{lyu2021conditional, luo2021score, qu2023conditional, melas2023pc2} also try to tackle the PC processing tasks mentioned in \sref{sec:four_tasks} by using different conditions to guide the PC generation. 
Despite previous efforts made some progress in leveraging diffusion models for point cloud completion \cite{lyu2021conditional}, denoising \cite{luo2021score}, upsampling \cite{qu2023conditional}, and single-image 3D reconstruction \cite{melas2023pc2}, no work has yet to integrate two modalities into a structured three-level-condition framework to guide the diffusion model in addressing the four tasks simultaneously.

%-------------------------------------------------------------------------
\vspace{-5pt}
\subsection{Image and Point Cloud Fusion}
\vspace{-5pt}
To effectively harness the inherent capabilities and potentials of both image and point cloud inputs, a robust fusion methodology becomes indispensable. 
The concept of image and point cloud fusion is rooted in the understanding that dense color or RGB information inherently carries 3D geometric details. As a result, images can be used as a reference to enhance the quality of the spatial point cloud. Stated by Huang et al. \cite{huang2022multi}, those fusion methods can be divided into the early-fusion \cite{xie2020pi,simon2019complexer,song2016deep}, the deep-fusion \cite{xu2018pointfusion,zhang2022maff,vora2020pointpainting}, and the late-fusion \cite{pang2020clocs, melotti2018multimodal, gu2019integrating}. Previous research on image and point cloud fusion has predominantly concentrated on high-level tasks \cite{song2016deep,vora2020pointpainting,xie2020pi,liang2022bevfusion,zhao2023attention} such as object detection and specific conventional tasks like point cloud completion and upsampling \cite{nguyen2022combination,zhang2021view,aiello2022cross}. 
However, these approaches are confined to a single task and limited scenarios.
% , which is a frequently overlooked potential in combining images and point clouds.
Hence, we propose SuperPC, a single diffusion model with a novel SMF strategy which combines the two modalities effectively and perform all the four tasks.

\vspace{-6pt}
\section{Background}
\label{sec:methodology}
\vspace{-5pt}

\begin{figure*}[t]
    \centering
    \includegraphics[width=1\linewidth]{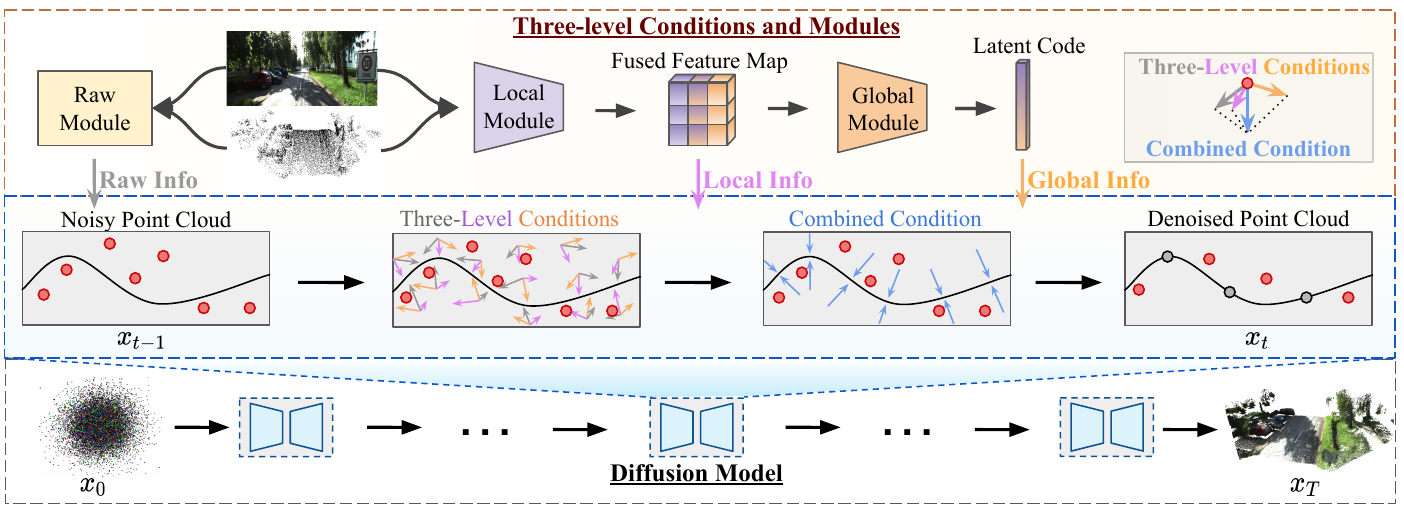} 
    % \caption{The main structure of the SuperPC model above shows that our method leverages the input image and point cloud to construct three-level conditions via the novel raw, local and global modules. These three-level conditions are fused into each step of the diffusion process to ensure SuperPC can leverage every level of information from the two input modalities to ensure the diffusion model can handle diverse of tasks and scenarios.}
    \vspace{-20pt}
    \caption{The architecture of the SuperPC model shown above integrates input images and point clouds to establish three-level conditions through innovative raw, local, and global modules. These conditions are seamlessly integrated into each step of the diffusion process, enabling SuperPC to utilize all levels of information from the two input modalities. }
    % This comprehensive approach ensures that the diffusion model is versatile, capable of handling a variety of tasks and scenarios effectively.}
    \label{fig:main_model}
    \vspace{-6pt}
\end{figure*}

\subsection{Conditional Diffusion Probabilistic Models} \label{sec:condition-diffusion}
\vspace{-5pt}
SuperPC employs Conditional Diffusion Probabilistic Models (CDPMs) \cite{saharia2022image, lugmayr2022repaint, luo2021diffusion, melas2023pc2, lyu2021conditional} to synthesize high-quality, dense point clouds, conditioned on the input images and point clouds. 
It consists of two Markov chains known as the forward processes and reverse processes \cite{ho2020denoising}.
% Both of the two sequences span a length of T. 
% In the SuperPC Model, we have designated T as 1000.

\paragraph{Forward Process} Also known as a Markov process that incrementally introduces Gaussian noises into the data distribution \(\boldsymbol{x_{0}}\), until transitioning it towards a pure Gaussian noise \(\boldsymbol{x_{T}}\). In other words, the point clouds progressively diffuse over time, finally forming purely chaotic points in a 3D space. 
This forward process from the target clean point cloud $\boldsymbol{x_0}$ to purely noisy point cloud $\boldsymbol{x_T}$  and and  can be formulated as the following probability distribution:

\vspace{-16pt}
\begin{equation}
\resizebox{0.72\linewidth}{!}{
    $
    \begin{aligned}
        q(\boldsymbol{x_{1:T}} | \boldsymbol{x_0}) &= \prod_{t=1}^{T} q(\boldsymbol{x_t} | \boldsymbol{x_{t-1}}), \\
        q(\boldsymbol{x_t | x_{t-1}}) &= \mathcal{N}\left(\boldsymbol{x_t}; \sqrt{1 - \beta_t} \boldsymbol{x_{t-1}}, \beta_t \mathbf{I}\right).
    \end{aligned}
    $
}
\end{equation}

The noise level is controlled by \(\beta_t\), and \(\boldsymbol{x_t}\) is sampled using $\boldsymbol{x_t} = \sqrt{\bar{\alpha}_t} \boldsymbol{x_0} + \sqrt{1 - \bar{\alpha}_t} \boldsymbol{\epsilon}$,
% \begin{equation} \label{eq:one_forward}
% \boldsymbol{x_t} = \sqrt{\bar{\alpha}_t} \boldsymbol{x_0} + \sqrt{1 - \bar{\alpha}_t} \boldsymbol{\epsilon},
% \end{equation}
where \(\boldsymbol{\epsilon}\) is standard Gaussian noise, $\alpha_t = 1 - \beta_t$ and $\bar{\alpha_t}=\prod_{s=1}^t \alpha_s$.

\paragraph{Reverse Process} This process is conditioned on the input point cloud \(P\) and image \(I\), and aims to remove the noise added by the forward process gradually as shown in the bottom \fref{fig:main_model}. 
This process is to predict the noise and reconstruct the point cloud data distribution $\boldsymbol{x_0}$ from \(\boldsymbol{x_T}\) based on the input point cloud \(P\) and image \(I\). 
% The process involves a multi-modal encoder \( \boldsymbol{\tau_\theta}(c,m) \) to obtain one of the conditions. 
% Another condition is the raw geometric information from the point cloud \(c\) input and the raw image features \(m\) introduced by the initialization layer.  
Mathematically, it can be expressed as the probability distribution \(p_{\boldsymbol{\theta}}\):

\vspace{-18pt}
\begin{equation}
\resizebox{0.88\linewidth}{!}{
    $
    \begin{aligned}
        p_{\boldsymbol{\theta}}(\boldsymbol{x_{1:T} | x_T}, P, I)
        &= \prod_{t=1}^{T} p_{\boldsymbol{\theta}}(\boldsymbol{x_{t-1}}|\boldsymbol{x_t}, P, I), \\
        \quad p_{\boldsymbol{\theta}}(\boldsymbol{x_{t-1}}|\boldsymbol{x_t}, P, I) 
        &= \mathcal{N}(\boldsymbol{x_{t-1}}; \boldsymbol{\mu_\theta}(\boldsymbol{x_t}, P, I, t), \sigma_t^2 \mathbf{I}),
    \end{aligned}
    $
}
\end{equation}
% where each step's distribution is:
% \begin{equation}
%     p_{\boldsymbol{\theta}}(\boldsymbol{x_{t-1}}|\boldsymbol{x_t}, P, I) 
%     = \mathcal{N}(\boldsymbol{x_{t-1}}; \boldsymbol{\mu_\theta}(\boldsymbol{x_t}, P, I, t), \sigma_t^2 \mathbf{I}),
% \end{equation}
where \(\boldsymbol{\mu_\theta}\) is the predicted mean of the Gaussian noise:
\begin{equation}
\resizebox{.328\vsize}{!}{
    $\boldsymbol{\mu_\theta}(\boldsymbol{x_t}, P, I, t)
    =\frac{1}{\sqrt{\alpha_t}} \left(\boldsymbol{x_t} - \frac{\beta_t}{\sqrt{1-\bar{\alpha_t}}}\boldsymbol{\epsilon_\theta}(\boldsymbol{x_t}, P, I, t)\right),$
    }
\end{equation}
where $\boldsymbol{\epsilon_\theta}$ is the main noise prediction network of SuperPC, which predicts the noise added in the forward process, guided by the three-level conditions in \sref{sec:three-level}.

\paragraph{Loss Function}
We aim to minimize the difference between the predicted and actual noise added during the forward process based on the input point cloud $P$ and image $I$. This is quantified using a mean squared error loss function $\mathcal{L}(\boldsymbol{\theta})$:
\begin{equation} \label{eq:loss}
\resizebox{0.88\linewidth}{!}{
    $
    \begin{aligned}
        \mathbb{E}_{\boldsymbol{x_0}, t, \boldsymbol{\epsilon}\sim\mathcal{N}(0, \mathbf{I})} 
        \left[ \|\boldsymbol{\epsilon} - \boldsymbol{\epsilon_\theta}(\sqrt{\bar{\alpha_t}}\boldsymbol{x_0} + \sqrt{1-\bar{\alpha_t}}\boldsymbol{\epsilon}, P, I, t)\|^2 \right],
    \end{aligned}
    $
}
\end{equation}
where \(\boldsymbol{\epsilon}\) is the actual Gaussian noise and \(\boldsymbol{\epsilon}_{\boldsymbol{\theta}}\) is the network predicted noise. This loss function ensures that the model refines its ability and knowledge to accurately generate the high-quality point cloud from the noised data while being guided by the three-level conditions.

%-------------------------------------------------------------------------
\vspace{-6pt}
\section{Methodology}
\label{sec:methodology2}
\vspace{-5pt}
As shown in \fref{fig:main_model}, we will explain the concept of the Three-Level-Conditioned (TLC) framework (middle of \fref{fig:main_model}, \sref{sec:three-level}), and details of three novel modules that apply the Spatial-Mix-Fusion (SMF) strategy to fuse the two input modalities and generate three-level conditions (top of \fref{fig:main_model}, \sref{sec:raw_module}, \sref{sec:local_module}, and \sref{sec:global_module}).

\vspace{-5pt}
\subsection{TLC Framework \& SMF Strategy} \label{sec:three-level}
\vspace{-5pt}
\paragraph{TLC Framework} 
First, we discuss the TLC framework, which integrates raw, local, and global-level abstracted information as conditions to guide the diffusion model in simultaneously handling four point cloud processing tasks. Our rationale is based on the observation that relying solely on a single type of condition can easily trap the model in a local minimum or cause it to overfit to one particular task.
For example, incorporating only global-level conditions into the diffusion model tends to produce basic shapes and lacks finer details, thereby hindering its capacity to generalize across more complex structures and expansive scenes \cite{luo2021diffusion}.
% For instance, the work \cite{luo2021diffusion} leverages object shape features as a global-level condition. 
% However, this method limits the model's capability to generate only simple shapes and lacks detail, which in turn obstructs its ability to generalize across complex structures and larger scenes.
To overcome this, we introduce three-level conditions into the diffusion model to enhance the model's performance and robustness to complex scenes and diverse tasks and get this new form of the loss function  $\mathcal{L}(\boldsymbol{\theta})$:
% Specifically, the  is redefined as:
\vspace{-5pt}
\begin{equation} \label{eq:loss2}
\resizebox{0.9\linewidth}{!}{
    $
    \begin{aligned}
        \mathbb{E}_{\boldsymbol{x_0}, \boldsymbol{\epsilon}, t}\left[ \|\boldsymbol{\epsilon} - \boldsymbol{\epsilon_\theta}(\sqrt{\bar{\alpha_t}}\boldsymbol{x_0} + \sqrt{1-\bar{\alpha_t}}\boldsymbol{\epsilon}, \boldsymbol{c_{\text{raw}}}, \boldsymbol{c_{\text{local}}}, \boldsymbol{c_{\text{global}}}, t)\|^2 \right],
    \end{aligned}
    $
}
\vspace{-5pt}
\end{equation}
where $\boldsymbol{c_{\text{raw}}}$, $\boldsymbol{c_{\text{local}}}$, and $\boldsymbol{c_{\text{global}}}$ represent the three-level conditions, collectively providing the combined instruction to guide the high-quality point clouds generating process.

\paragraph{SMF Strategy} 
Previous works, whether employing early fusion \cite{xie2020pi,simon2019complexer}, deep fusion \cite{xu2018pointfusion,zhang2022maff}, or late fusion \cite{pang2020clocs, melotti2018multimodal}, have demonstrated strong performance in 3D object detection. 
However, we observed that these strategies are less effective for point cloud processing tasks. Specifically, late fusion in 3D object detection combines detection results from the image and point cloud branches at the final stage, making it unsuitable for point cloud processing tasks. 
Furthermore, as shown in our experiments (Appendix \tref{AblationResults}), both early and late fusion strategies show poor effectiveness across all three benchmarks. To address this challenge, we introduce the Spatial-Mix-Fusion (SMF) strategy, a loose coupling of early and deep fusion approaches.
Specifically, we designed a \textbf{dual-spatial early fusion} and an \textbf{attention-based deep fusion} to integrate image and point cloud modalities into the diffusion conditions, both leveraging the spatial features of the partially denoised point cloud in the main diffusion network.
We next delve into the specifics of constructing the TLC framework and the SMF strategy with the global, local, and raw modules.

% To make it easier to understand, we will first introduce the mechanism of conditional diffusion models in \sref{sec:condition-diffusion}.
% This will be followed by the details of the three-level-conditioned framework and its three significant modules, namely a raw, a local, and a global module to generate the three-level conditions in \sref{sec:three-level}.

% \begin{figure}[t]
%     \begin{subfigure}{.41\textwidth}
%       \centering
%       % include first image
%       \includegraphics[width=1\linewidth]{images/raw_module.pdf}  
%       \caption{Raw module.}
%       \label{fig:raw_module}
%     \end{subfigure}
%     \begin{subfigure}{.55\textwidth}
%       \centering
%       % include second image
%       \includegraphics[width=1\linewidth]{images/feature_fusion_module.pdf}  
%       \caption{Local and global modules.}
%       \label{fig:local_global_module}
%     \end{subfigure}
%     \vspace{10pt}
%     \caption{The details of the three-level modules.}
%     \label{fig:three-modules}
% \end{figure}

\begin{figure*}[t]
    \centering
    \includegraphics[width=1\linewidth]{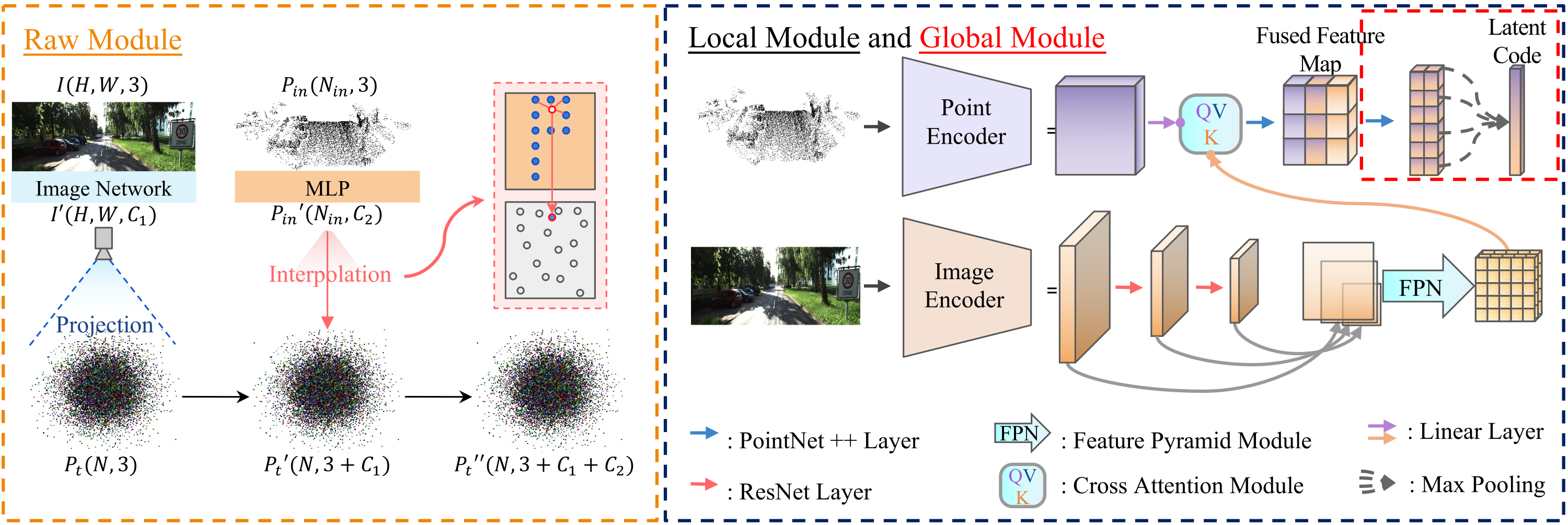} 
    \vspace{-20pt}
    \caption{On the left, the raw module integrates the raw information of the image and point cloud into the target point cloud as the raw-level condition via the image projection and the point interpolation. On the right, the local module encodes the two inputs into feature maps, which are then fused using cross-attention to produce a local fused feature map as the local-level condition. 
    Next, the global module condenses this feature map into a global latent code as the global-level condition.}
    \label{fig:three-modules}
    \vspace{-10pt} 
\end{figure*}

\vspace{-5pt}
\subsection{Raw Module \& Dual-Spatial Early Fusion} \label{sec:raw_module}
\vspace{-5pt}
The raw module shown in \fref{fig:three-modules} (left) aims to preserve and fuse detailed texture and spatial information.
It can be divided into two components: the image features projection and the point cloud features interpolation, which realize the dual-spatial early fusion.
The two components are to incorporate and fuse the raw sensor information into each diffusion step to guide the target point cloud generating process, respectively.
Specifically, the image branch utilizes the rasterization technique described in \cite{ravi2020accelerating, melas2023pc2} to project image features onto the partially denoised point cloud, while the point cloud branch uses spatial interpolation technique to align the input point cloud with the partially denoised point cloud in the main diffusion network.

\vspace{-1pt}
\paragraph{Image Features Projection} 
Before projecting the image to the point cloud, a normal 2D image network (like ResNet \cite{shinohara2021point2color} or ViT \cite{dosovitskiy2020image}) is used to increase the image feature dimension from $I(H, W, 3)$ to $I'(H, W, C_1)$, providing richer raw RGB information. Then the computationally efficient {\fontfamily{qcr}\selectfont PointRasterizer} class of PyTorch3D is used to project the dense image features to the point cloud at the beginning of each diffusion step. 
As a result, the point cloud dimension is increased from $P_t(N,3)$ to $P_t'(N,3+C_1)$.
This step is crucial for roughly aligning the images and point clouds in terms of both raw spatial structural information and texture color information.

\vspace{-1pt}
\paragraph{Point Cloud Features Interpolation}
Subsequently,  we use a special spatial interpolation method to align the raw features of the input point cloud \( P_{in}(N_{in},3) \) with the image-aligned features of \( P_t'(N,3+C_1) \).
Specifically, we first employ a multilayer perceptron (MLP) to increase the feature dimension of the input point cloud from \( P_{in}(N_{in},3) \) to $P_{in}'(N_{in}, C_2)$.
Due to the disorder of the point cloud, we aggregated their features using the $k$-nearest neighbors based on the inverse-distance-weighted average algorithm \cite{qi2017pointnet++}.
% a k-nearest neighbor (KNN) is used to find the $k$ nearest points in $P_{in}'$ for each point in $P_t'$. Next, the inverse-distance-weighted average \cite{qi2017pointnet++} is used to calculate the features added to $P_t'$. 
Assuming the weighted features of the points in $P_t'$ are $\boldsymbol{f^{(j)}} (j = 1, 2, ..., N)$ and features of the $k$ (here we use $k=4$) nearest points in $P_{in}'$ are $\boldsymbol{f_i^{(j)}} (i = 1, 2, ..., k)$, we can write the aggregated point features as:

\begin{equation} 
\resizebox{0.68\linewidth}{!}{
    $
    \begin{aligned}
        &\boldsymbol{f^{(j)}} 
        = \nicefrac{\sum_{i=1}^k w(\boldsymbol{x_j}, \boldsymbol{x_i}) \boldsymbol{f_i^{(j)}}}{\sum_{i=1}^k w(\boldsymbol{x_j}, \boldsymbol{x_i})}, \\
        &w(\boldsymbol{x_j}, \boldsymbol{x_i}) 
        = \nicefrac{1}{d(\boldsymbol{x_j}, \boldsymbol{x_i})^2},
    \end{aligned}
    $
}
\end{equation}
% \begin{equation}
% \resizebox{0.88\linewidth}{!}{
%     $
%     \begin{aligned}
%         p_{\boldsymbol{\theta}}(\boldsymbol{x_{1:T} | x_T}, P, I)
%         &= \prod_{t=1}^{T} p_{\boldsymbol{\theta}}(\boldsymbol{x_{t-1}}|\boldsymbol{x_t}, P, I), \\
%         \quad p_{\boldsymbol{\theta}}(\boldsymbol{x_{t-1}}|\boldsymbol{x_t}, P, I) 
%         &= \mathcal{N}(\boldsymbol{x_{t-1}}; \boldsymbol{\mu_\theta}(\boldsymbol{x_t}, P, I, t), \sigma_t^2 \mathbf{I}).
%     \end{aligned}
%     $
% }
% \end{equation}
where the $d(\boldsymbol{x_j}, \boldsymbol{x_i})$ in the second equation indicates the Euclidean distance between points $\boldsymbol{x_j}$ and $\boldsymbol{x_i}$. 

\vspace{-1pt}
\paragraph{Dual-Spatial Early Fusion}
The two components described above result in a point cloud $P_t''(N, 3+C_1+C_2)$ which successfully fuse the two modalities' raw-level information to the main diffusion network via the dual-spatial operations - (1) the spatial interpolation and (2) the point-rasterizer projection based on the spatial positions of the partially-denoised point cloud of the diffusion network.

\vspace{-5pt}
\subsection{Local Module \& Attention-Based Deep Fusion} \label{sec:local_module}
\vspace{-5pt}
The local module shown in \fref{fig:three-modules} (right) is designed to introduce local object-level information to guide the diffusion model.
Specifically, the local module merges two modalities into a fused feature map, which is then integrated into the main diffusion network as the local condition.
% Unlike the standalone raw module, the other two modules—the local and global—are interconnected as depicted in right side of \fref{fig:three-modules}.
% Here, the local module takes the single image and point cloud as inputs and fuses them into a local feature map and then the global module will compress the feature map into a one-dimensional latent code.
% This section delves into the details of the two modules - the input encoding and feature fusion of the local module, and the feature compression of the global module.

\captionsetup[subfigure]{aboveskip=1pt}
\begin{figure*}[t]
    \begin{subfigure}{.346\textwidth}
      \centering
      % include first image
      \includegraphics[width=1\linewidth]{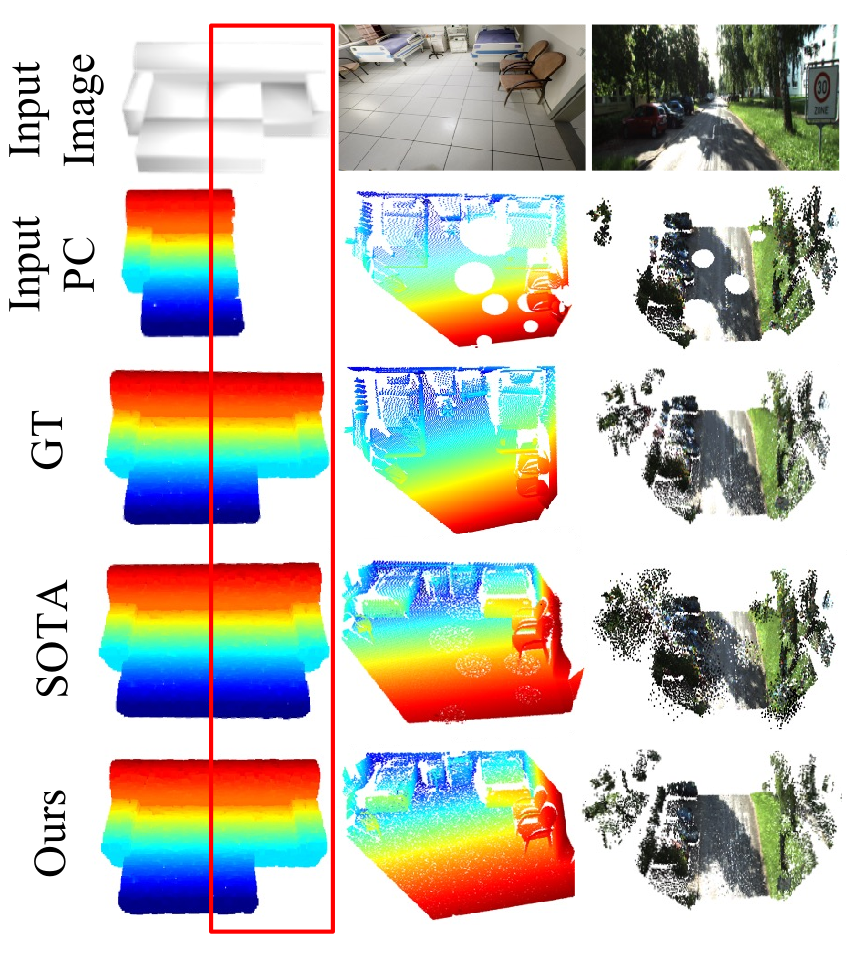}  
      \caption{Completion.}
      \label{fig:completion}
    \end{subfigure}
    \begin{subfigure}{.328\textwidth}
      \centering
      % include second image
      \includegraphics[width=1\linewidth]{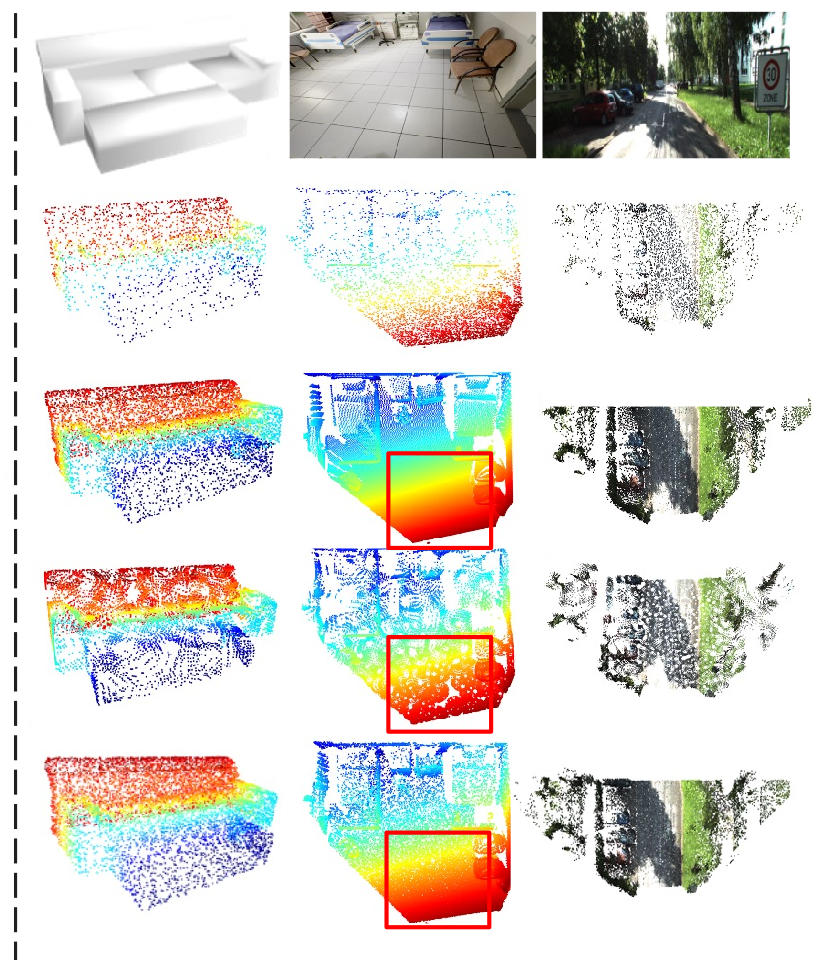}  
      \caption{Upsampling.}
      \label{fig:upsampling}
    \end{subfigure}
    \begin{subfigure}{.316\textwidth}
      \centering
      % include third image
      \includegraphics[width=1\linewidth]{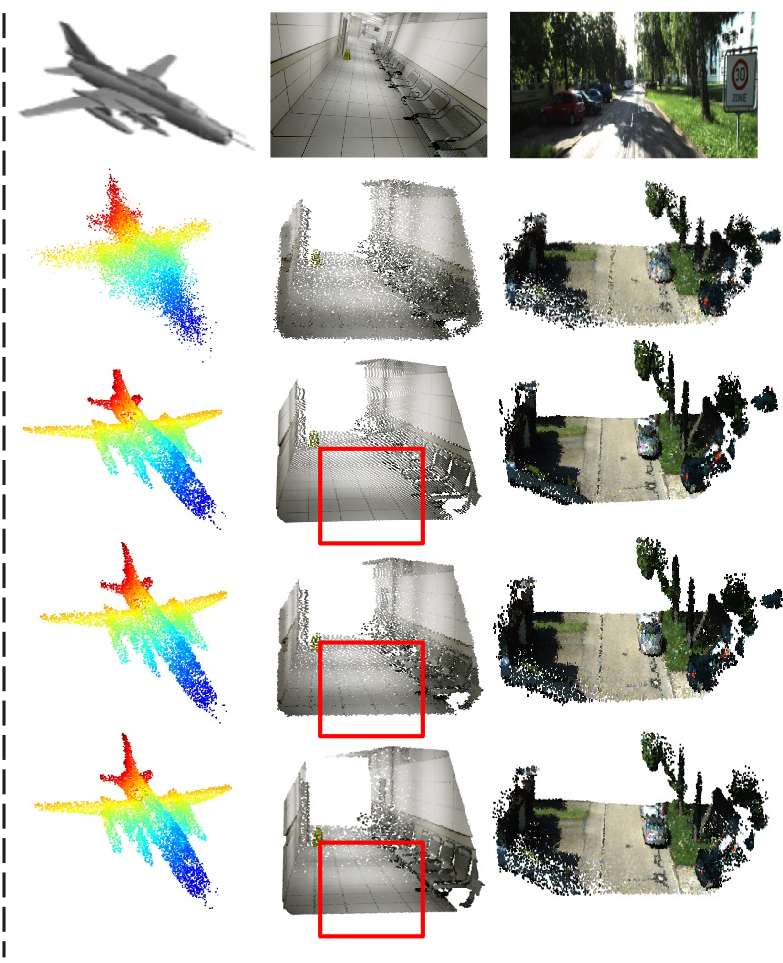}  
      \caption{Denoising.}
      \label{fig:denoising}
    \end{subfigure}
    \vspace{-18pt}
    \caption{The qualitative results on the point cloud (a) completion, (b) upsampling, and (c) denoising tasks. For each subfigure, from left to right are the results for ShapeNet, TartanAir, and KITTI-360. Larger figures and more qualitative results are presented in Appendix \ref{sec:more_quantitative}.}
    \label{fig:LLM-enabled-IL-structure}
    \vspace{-10pt}
\end{figure*}

\vspace{-1pt}
\paragraph{Inputs Encoding}
The local module uses the modality-specific encoders to extract features from the two inputs and fuse them into a local feature map based on cross-attention.
Specifically, we use PointNet++ \cite{qi2017pointnet++} and ResNet \cite{shinohara2021point2color} as the point cloud and image encoders, respectively.
% The first step - encoding of input point clouds and images is handled by two distinct neural networks: the point encoder (PointNet++ \cite{qi2017pointnet++} or Point-voxel CNN \cite{liu2019point} backbone) and the image encoder (ResNet \cite{shinohara2021point2color}), respectively. 
% The hierarchical feature learning architecture of the backbones empower the point encoder to extract the local features from these unstructured, irregular point cloud data. 
We leverage the Feature Pyramid Networks (FPN) to map multi-scale spatial hierarchies to one comprehensive feature map for local representation. 
This is essential for capturing multi-scale semantic features, ensuring a detailed and multi-dimensional representation of the input visual information.

\vspace{-1pt}
\paragraph{Attention-Based Deep Fusion}
Following the separate features extracting, the local module employs the cross-attention mechanism to fuse the images and point clouds. This fusion strategy is adapted from generative models \cite{rombach2022high, shen2024icafusion, wei2020multi}, which are particularly beneficial for handling diverse types of conditioning inputs.
% The process begins by transforming the conditioning inputs from both modalities into intermediate representations. T
% he image input \( y \) is converted into a feature representation \(\tau_{\theta}(y)\) in \(\mathbb{R}^{M \times d_{\tau}}\). 
The image features and point cloud features are transformed into a compatible feature space.
The cross-attention mechanism is then applied to these intermediate representations, which can be formulated as: $\text{Attention}(Q, K, V) = \text{softmax}(\nicefrac{QK}{\sqrt{T}}) \cdot V$, where \( Q \) represents the query tensor derived from image encoder's intermediate representation of the image feature map, while \( K \) and \( V \) are the key and value tensors obtained from the point encoder. These tensors are processed through learnable projection matrices \( W_Q(f_{img}) \), \( W_K(f_{pc}) \), and \( W_V(f_{pc}) \), which are passing the outputs of the image and the point encoders through three different linear layers to obtain Q, K and V, respectively. This cross-attention block aligns and integrates the point cloud spatial features with the multi-scale image semantic features, creating a unified fused feature map encoding both local spatial and visual features of the point cloud and the image. 
% The fused feature map is in the form of deep points \( P_f'(N_f,3+C_f) \) which will then be integrated into the main diffusion network via the spatial interpolation discussed in \sref{sec:raw_module}.

\vspace{-5pt}
\subsection{Global Module} \label{sec:global_module}
\vspace{-5pt}
% \paragraph{Global Module - Feature Compression}
Subsequently, a one-dimensional latent code representing global structural features is generated through a simple PointNet++ layer and a max pooling layer, a technique for emphasizing high-level features while reducing dimensionality. 
These two modules enable a seamless fusion of image and point cloud data, ensuring that the generated output effectively encapsulates the characteristics of both modalities. 
Consequently, they guide the diffusion process to extract and learn the structure details and the semantic consistency of the point clouds and images.
We have included details of the implementation of the three modules in Appendix \ref{sec:detail_structure}.

\vspace{-6pt}
\section{Experiments}
\label{sec:experiments}
\vspace{-5pt}

\paragraph{Benchmarks and Datasets} \label{sec:datasets}
Previous methods have mainly reported performance on the ShapeNet dataset \cite{chang2015shapenet}.
However, it only contains simple 3D objects like chairs and planes without color, limiting its ability to evaluate like in scene-level scenarios and point cloud colorization.
% It can reflect the performance on point cloud completion and denoising, but it only contains simple 3D objects such as chairs and planes without color, which limits its applications such as point cloud colorization.
Therefore, We have designed three new benchmarks for point cloud processing tasks: one object-level (based on ShapeNet \cite{chang2015shapenet} dataset) and two scene-level benchmarks (based on TartanAir \cite{wang2020tartanair} and KITTI360 \cite{liao2022kitti} datasets). Additionally, these benchmarks are organized into two evaluation tracks to assess generalization ability—Object-to-Scene and Sim-to-Real.
% To demonstrate the generalization ability, tackle the object-to-scene and sim-to-real gaps, and fully test performance on all four tasks, we introduce two more datasets, i.e., TartanAir \cite{wang2020tartanair} and KITTI-360 \cite{liao2022kitti} to cover more complicated and larger scale scenes in simulation and real-world, respectively.
The detailed setup of the three benchmarks is presented in Appendix \ref{sec:dataset_details}.                                          
%-------------------------------------------------------------------------

\vspace{-1pt}
\paragraph{Evaluation Metrics} \label{sec:metrics}
% Given that our goal is to simultaneously tackle four tasks using a single model, the evaluation metrics must be applicable across all tasks. 
For comprehensiveness, we select the most widely used metrics for point cloud processing including Earth Mover's Distance (EMD) and F1.
Additionally, we also use Density-aware Chamfer Distance (DCD) \cite{wu2021density} instead of the traditional Chamfer Distance (CD) as CD is sensitive to mismatched local density \cite{wu2021density}.
% The three most widely used metrics are Chamfer Distance (CD), Earth Mover's Distance (EMD), and F1.
% In this work, we adopt EMD, F1, and the Density-aware Chamfer Distance (DCD) \cite{wu2021density} instead of the traditional CD as CD is sensitive to mismatched local density. 
% Our analysis, as illustrated in \fref{fig:upsampling}, highlights that local density is a critical aspect of evaluating point cloud processing methods, as certain approaches may fail to work and result in unevenpoint density distribution.
The specifics of the three metrics are detailed in Appendix \ref{sec:metrics_details}.

We next evaluate SuperPC against SOTA models across all four individual tasks and the combined task (\sref{sec:SOTAs_comparison}), assess its generalization through object-to-scene and sim-to-real tests (\sref{sec:Generalization_experiments}).

%-------------------------------------------------------------------------
\vspace{-5pt}
\subsection{Performance} \label{sec:SOTAs_comparison}
\vspace{-5pt}
\setlength\heavyrulewidth{0.36ex}
\begin{table*}[ht]
\caption{Results compared with SOTA models on point cloud completion, upsampling, denoising, and combination tasks on the ShapeNet, TartanAir, and KITTI-360 benchmarks (from top to bottom). The cross mark ``\protect\crossmark'' indicates that the model is incapable of performing the specified task. Different SOTA combinations, e.g., ``\cite{he2023grad}$\rightarrow$\cite{nunes2024scaling}$\rightarrow$\cite{luo2021score}'', are selected for different benchmarks because they are the top performers for that specific category. A more detailed comparison of these SOTA combinations can be found in Appendix  \ref{sec:different_integration}.}
\label{SOTAResults}
\centering
% ShapeNet
\scalebox{0.8}{
\begin{tabular}{cccccccccccccc}
\toprule
\multirow{3}{*}{\textbf{ShapeNet} \cite{chang2015shapenet}} & \multicolumn{3}{c}{Completion} & \multicolumn{3}{c}{Upsampling} & \multicolumn{3}{c}{Denoising} & \multicolumn{3}{c}{Combination} \\ \cmidrule(lr){2-4} \cmidrule(lr){5-7} \cmidrule(lr){8-10} \cmidrule(lr){11-13}
                    & DCD ($\downarrow$)         & EMD ($\downarrow$)       & F1 ($\uparrow$)       & DCD ($\downarrow$)          & EMD ($\downarrow$)       & F1 ($\uparrow$)       & DCD ($\downarrow$)       & EMD ($\downarrow$)       & F1 ($\uparrow$)       & DCD ($\downarrow$)       & EMD ($\downarrow$)       & F1 ($\uparrow$)\\ \midrule
AdaPoinTr \cite{yu2023adapointr}             & \underline{0.462}        & \underline{2.12}        & \underline{0.423}        & $\crossmark$            & $\crossmark$            & $\crossmark$            & 0.562            & 3.61           & 0.405            & $\crossmark$            & $\crossmark$           & $\crossmark$            \\

GradPU \cite{he2023grad}                & $\crossmark$            & $\crossmark$           & $\crossmark$            & 0.298        & 1.27         & 0.589        & 0.533            & 3.28           & 0.412            & $\crossmark$            & $\crossmark$           & $\crossmark$            \\

DDPMPU \cite{qu2024conditional}                & $\crossmark$            & $\crossmark$           & $\crossmark$            & \textbf{0.281}        & \textbf{1.13}         & \textbf{0.659}        & 0.313            & 1.37           & 0.769            & $\crossmark$            & $\crossmark$           & $\crossmark$            \\

ScoreDenoise \cite{luo2021score}           & $\crossmark$            & $\crossmark$           & $\crossmark$            & 0.346        & 1.65         & 0.537        & 0.291        &  1.21       & 0.812        & $\crossmark$            & $\crossmark$           & $\crossmark$            \\ 

PD-LTS \cite{Mao_2024_CVPR}           & $\crossmark$            & $\crossmark$           & $\crossmark$            & $\crossmark$            & $\crossmark$           & $\crossmark$        & \textbf{0.280}        &  \textbf{1.12}       & \textbf{0.856}        & $\crossmark$            & $\crossmark$           & $\crossmark$            \\ 

LiDiff \cite{nunes2024scaling}                 & 0.484        & 2.43        & 0.401        & 0.312        & 1.38         & 0.526        & $\crossmark$            & $\crossmark$           & $\crossmark$            & $\crossmark$            & $\crossmark$           & $\crossmark$            \\ 

\cite{Mao_2024_CVPR}$\rightarrow$\cite{yu2023adapointr}$\rightarrow$\cite{qu2024conditional}   
                               & 0.462        & 2.12        & 0.423        & 0.281        & 1.13         & 0.659        & 0.280        &  1.12       & 0.856        & 0.509        & 2.71        & 0.382\\ 
SuperPC (ours)                &\textbf{0.387}&\textbf{1.67}&\textbf{0.557}&\underline{0.293}&\underline{1.25} &\underline{0.631}&\underline{0.285}&\underline{1.16}&\underline{0.837}&\textbf{0.476}&\textbf{2.21}&\textbf{0.409}\\ \bottomrule
\end{tabular}
}
% TartanAir
\scalebox{0.8}{
\begin{tabular}{cccccccccccccc}
\toprule
\multirow{3}{*}{\textbf{TartanAir} \cite{wang2020tartanair}} & \multicolumn{3}{c}{Completion} & \multicolumn{3}{c}{Upsampling} & \multicolumn{3}{c}{Denoising} & \multicolumn{3}{c}{Combination} \\ \cmidrule(lr){2-4} \cmidrule(lr){5-7} \cmidrule(lr){8-10} \cmidrule(lr){11-13}
                    & DCD ($\downarrow$)         & EMD ($\downarrow$)       & F1 ($\uparrow$)       & DCD ($\downarrow$)          & EMD ($\downarrow$)       & F1 ($\uparrow$)       & DCD ($\downarrow$)       & EMD ($\downarrow$)       & F1 ($\uparrow$)       & DCD ($\downarrow$)       & EMD ($\downarrow$)       & F1 ($\uparrow$)\\ \midrule
AdaPoinTr \cite{yu2023adapointr}             & 0.573        & 3.59        & 0.349        & $\crossmark$            & $\crossmark$            & $\crossmark$            & 0.604            & 3.91           & 0.229            & $\crossmark$            & $\crossmark$           & $\crossmark$            \\

GradPU \cite{he2023grad}                & $\crossmark$            & $\crossmark$           & $\crossmark$            & \underline{0.527}        & \underline{3.31}         & \underline{0.313}        & 0.572            & 3.85           & 0.235            & $\crossmark$            & $\crossmark$           & $\crossmark$            \\

DDPMPU \cite{qu2024conditional}                & $\crossmark$            & $\crossmark$           & $\crossmark$            & 0.541        & 3.49         & 0.302        & 0.389            & 1.91           & 0.438            & $\crossmark$            & $\crossmark$           & $\crossmark$            \\

ScoreDenoise \cite{luo2021score}           & $\crossmark$            & $\crossmark$           & $\crossmark$            & 0.617        & 3.96         & 0.224        & \underline{0.346}        &  \underline{1.65}       & \underline{0.473}        & $\crossmark$            & $\crossmark$           & $\crossmark$            \\ 

PD-LTS \cite{Mao_2024_CVPR}           & $\crossmark$            & $\crossmark$           & $\crossmark$            & $\crossmark$            & $\crossmark$           & $\crossmark$        & 0.368        &  1.89       & 0.451        & $\crossmark$            & $\crossmark$           & $\crossmark$            \\ 

LiDiff \cite{nunes2024scaling}                 & \underline{0.559}        & \underline{3.51}        & \underline{0.356}        & 0.562        & 3.62         & 0.248        & $\crossmark$            & $\crossmark$           & $\crossmark$            & $\crossmark$            & $\crossmark$           & $\crossmark$            \\ 

\cite{he2023grad}$\rightarrow$\cite{nunes2024scaling}$\rightarrow$\cite{luo2021score}   
                               & 0.559        & 3.51        & 0.356        & 0.527        & 3.31         & 0.313        & 0.346        &  1.65       & 0.473        & 0.583        & 3.64        & 0.369 \\ 
SuperPC (ours)                &\textbf{0.538}&\textbf{3.46}&\textbf{0.363}&\textbf{0.492}&\textbf{2.98} &\textbf{0.356}&\textbf{0.298}&\textbf{1.38}&\textbf{0.631}&\textbf{0.558}&\textbf{3.53}&\textbf{0.384}\\ \bottomrule
\end{tabular}
}
% KITTI-360
\scalebox{0.8}{
\begin{tabular}{cccccccccccccc}
\toprule
\multirow{3}{*}{\textbf{KITTI-360} \cite{liao2022kitti}} & \multicolumn{3}{c}{Completion} & \multicolumn{3}{c}{Upsampling} & \multicolumn{3}{c}{Denoising} & \multicolumn{3}{c}{Combination} \\ \cmidrule(lr){2-4} \cmidrule(lr){5-7} \cmidrule(lr){8-10} \cmidrule(lr){11-13}
                    & DCD ($\downarrow$)         & EMD ($\downarrow$)       & F1 ($\uparrow$)       & DCD ($\downarrow$)          & EMD ($\downarrow$)       & F1 ($\uparrow$)       & DCD ($\downarrow$)       & EMD ($\downarrow$)       & F1 ($\uparrow$)       & DCD ($\downarrow$)       & EMD ($\downarrow$)       & F1 ($\uparrow$)\\ \midrule
AdaPoinTr \cite{yu2023adapointr}             & 0.663        & 9.27        & 0.293        & $\crossmark$            & $\crossmark$            & $\crossmark$            & 0.621            & 7.85           & 0.329            & $\crossmark$            & $\crossmark$           & $\crossmark$            \\

GradPU \cite{he2023grad}                & $\crossmark$            & $\crossmark$           & $\crossmark$            & \underline{0.597}        & \underline{6.92}         & \underline{0.354}        & 0.589            & 6.47           & 0.368            & $\crossmark$            & $\crossmark$           & $\crossmark$            \\

DDPMPU \cite{qu2024conditional}                & $\crossmark$            & $\crossmark$           & $\crossmark$            & 0.601        & 7.18         & 0.348        & 0.407            & 3.95           & 0.542            & $\crossmark$            & $\crossmark$           & $\crossmark$            \\

ScoreDenoise \cite{luo2021score}           & $\crossmark$            & $\crossmark$           & $\crossmark$            & 0.718        & 9.93         & 0.254        & \underline{0.369}        &  \underline{3.21}       & \underline{0.604}        & $\crossmark$            & $\crossmark$           & $\crossmark$            \\ 

PD-LTS \cite{Mao_2024_CVPR}           & $\crossmark$            & $\crossmark$           & $\crossmark$            & $\crossmark$            & $\crossmark$           & $\crossmark$        & 0.381        & 3.37       & 0.575        & $\crossmark$            & $\crossmark$           & $\crossmark$            \\ 

LiDiff \cite{nunes2024scaling}                 & \underline{0.649}        & \underline{9.03}        & \underline{0.307}        & 0.609        & 7.35         & 0.341        & $\crossmark$            & $\crossmark$           & $\crossmark$            & $\crossmark$            & $\crossmark$           & $\crossmark$            \\ 

\cite{he2023grad}$\rightarrow$\cite{nunes2024scaling}$\rightarrow$\cite{luo2021score}   
                               & 0.649        & 9.03        & 0.307        & 0.597        & 6.92         & 0.354        & 0.369        & 3.21        & 0.604        & 0.725        & 10.06       & 0.347 \\ 
SuperPC (ours)                &\textbf{0.632}&\textbf{8.82}&\textbf{0.324}&\textbf{0.577}&\textbf{6.73} &\textbf{0.369}&\textbf{0.327}&\textbf{2.86}&\textbf{0.615}&\textbf{0.681}&\textbf{9.58}&\textbf{0.365}\\ \bottomrule
\end{tabular}
}
\vspace{-6pt}
\end{table*}

%%%%%%%%%%%%%%%%%%%%%%%%%%%%%%%%%%%%%%%%%%%%%%%%%%%%%%%%%%%%%%%%%%%%%%%%%%%%%%%%%%%%%%%%%%%%%%%%%%

\label{sec:sotas_experiments}
% We take the SOTA models in point cloud completion (PC \cite{yu2023adapointr}), point cloud upsampling (PU \cite{he2023grad}), and point cloud denosing (PD \cite{luo2021score}), as well as their combined models as our baselines.
% We take the most well-known and recent point cloud processing models \cite{yu2023adapointr,he2023grad,luo2021score, nunes2024scaling} (each model we select is the SOTA for one individual task), as well as their combined models as our baselines.
We select the most well-known and recent point cloud processing models \cite{yu2023adapointr,he2023grad,luo2021score, nunes2024scaling, qu2024conditional, Mao_2024_CVPR} as well as their combined models to serve as our baselines (all the chosen models are the SOTAs for the individual tasks).
% Since widely accepted point cloud colorization metrics are unavailable and the superiority of our SuperPC can be easily visualized, we will compare SuperPC with the baseline learning method \cite{liu2019pccn} and the SOTA projection method, showing qualitative results.
% As we propose utilizing a single model to simultaneously address multiple tasks, our experiments include comparisons not only with baselines of four separate point cloud processing tasks but also with combined tasks against sequences of state-of-the-art (SOTA) point cloud processing methods. 
% We present our experiments in a sequence that prioritizes quantitative assessments followed by qualitative evaluations. 
% In this section, we first discuss the tasks of completion, upsampling, and denoising, as well as the combination tasks that can be quantitatively evaluated. 
% In these three tasks, we conducted experimental comparisons of our method with the SOTA models for each task across three datasets. 
%
% These SOTA models are AdaPtr \cite{yu2023adapointr} for point cloud completion, Grad-PU \cite{he2023grad} for point cloud upsampling, and score-based point cloud denoising \cite{luo2021score}. 
All models were subjected to identical training-validation-testing settings. 
% To fulfill the training and validation requirements of all the models being tested, the point clouds from both the object-level and scene-level datasets were sampled to consist of 8,192 and 46,080 points, respectively.

As can be seen in \tref{SOTAResults}, none of the baseline models can conduct all the tasks. Therefore, we integrate the best model for each task together to tackle the combination task and compare the integrated method with SuperPC. 
How we evaluate the best integration model and additional qualitative results can be found in Appendix \ref{sec:different_integration} and \ref{sec:more_quantitative}.
Remarkably, SuperPC surpasses all the models in all three tasks and their combination task across the three benchmarks. Besides the general quantitative results, we next delve into the details and qualitative results of each task.

\paragraph{Point Cloud Completion (PC)}
For the completion task, we follow the moderate sampling setting in \cite{yu2023adapointr} on the ShapeNet dataset, i.e., 50\% of the points are used as the incompleted input point clouds. 
On the two scene-level benchmarks, we develop a simple random patch removal algorithm to generate point clouds with random blank areas as shown in \fref{fig:completion} and \ref{fig:combination}. 
% As shown in Table \ref{SOTAResults}, SuperPC outperforms the all the models on all three datasets.
% Notably, the SOTA method for completion tasks shows commendable performance on scene datasets, a distinction from other tasks. The reason might be they utilize not only the global but also the local information in their structure \cite{yu2023adapointr}. 
Notably, as illustrated in the first row of \fref{fig:completion}, the best baseline method \cite{yu2023adapointr} inaccurately generates a completed symmetrical sofa, diverging from the actual asymmetrical shape of the ground truth. In contrast, SuperPC integrates information from the image input to accurately complete the asymmetrical shape.

\paragraph{Point Cloud Upsampling (PU)}
For this task, we set the integer testing upsampling rate as $\times8$ for all three benchmarks to meet the inference limitation on the integer upsampling rate. Unlike the majority of previous upsampling methods \cite{yu2018pu, yifan2019patch, li2019pu}, our model can take an arbitrary real number upsampling ratio like $1.3425$. 
% Crucially, our method surpasses the SOTA performance across all three datasets. 
Moreover, \fref{fig:upsampling} reveals that the point clouds generated by the SOTA \cite{he2023grad} exhibit an odd pattern of point concentration across all benchmarks. 
% That phenomenon cannot be adequately assessed using the Chamfer Distance, as discussed in \sref{sec:DCD}. 
Noticeably, SuperPC surpasses the SOTA and baseline methods on both scene-level benchmarks, while DDPMPU \cite{qu2024conditional} shows only marginally better performance on the object-level benchmark.
Unlike previous works, SuperPC can produce not only precise but also evenly distributed point clouds with an arbitrary upsampling rate and shows dominating performance on the scene-level benchmarks.

\begin{figure*}[ht]
  \begin{minipage}[b]{.67\linewidth}
    \centering
    \scalebox{0.83}{
        \begin{tabular}{ccccccccc}
        \toprule
        \multirow{3}{*}{Tasks} & \multirow{3}{*}{Method} & \multicolumn{3}{c}{Object-to-Scene} & \multicolumn{3}{c}{Sim-to-Real} \\ \cmidrule(lr){3-5} \cmidrule(lr){6-8}  
                              &                         & DCD ($\downarrow$)          & EMD ($\downarrow$)       & F1 ($\uparrow$)       & DCD ($\downarrow$)       & EMD ($\downarrow$)       & F1 ($\uparrow$)      \\ \midrule
        \multirow{2}{*}{Completion}  & PC           & 0.901        & 29.8         & 0.122        & 0.873        & 19.39        & 0.156       \\
                                    & SuperPC (ours)      &\textbf{0.857}&\textbf{24.2} &\textbf{0.159}&\textbf{0.789}&\textbf{13.93} &\textbf{0.225}\\ \midrule
        \multirow{2}{*}{Upsampling}  & PU           &\textbf{0.613}&\textbf{6.81} &\textbf{0.316}& 0.693        & 10.42        & 0.279        \\
                                    & SuperPC (ours)      & 0.738        & 11.4         & 0.243         &\textbf{0.648}&\textbf{8.96} &\textbf{0.336}\\ \midrule
        \multirow{2}{*}{Denoising}   & PD           & 0.368        & 3.67         & 0.431        & 0.588        &  3.42        & 0.343          \\ 
                                    & SuperPC (ours)      &\textbf{0.333}&\textbf{2.91} &\textbf{0.513}&\textbf{0.552}&\textbf{2.98} &\textbf{0.458}\\ \midrule
        \multirow{2}{*}{Combination} 
        & PU$\rightarrow$PC$\rightarrow$PD                 & 0.917        & 31.3         & 0.116        & 0.891        & 21.45        & 0.139            \\ 
        & SuperPC (ours)                                  &\textbf{0.861}&\textbf{25.8} &\textbf{0.143}&\textbf{0.802}&\textbf{14.86}&\textbf{0.214}\\ \bottomrule
        \end{tabular}
    }
    \captionsetup{width=.66\linewidth}
    \captionof{table}{Generalization ability experiment on the four point cloud processing tasks. (5\% data used for fine-tuning.)}
    \label{GenResults}
  \end{minipage}
  \begin{minipage}[b]{.33\linewidth}
    \centering
    \includegraphics[width=0.94\linewidth]{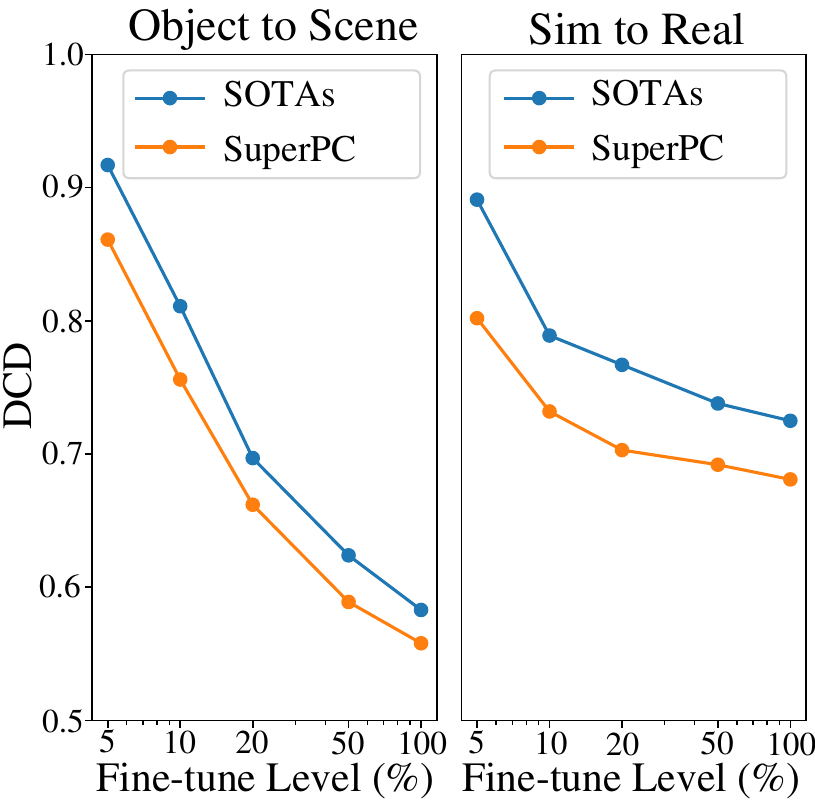}
    \vspace{-6pt}
    \captionsetup{width=.82\linewidth}
    \captionof{figure}{Combination task.}
    \label{fig:generalization_test}
  \end{minipage}\hfill
  \vspace{-22pt}
\end{figure*}

\vspace{-2pt}
\paragraph{Point Cloud Denoising (PD)}
In the denoising task, the accurate ground truth point clouds are perturbed by Gaussian noise with standard deviation from $0.5\%$ to $2\%$ of the bounding sphere radius to generate the input noisy data following the setting in \cite{luo2021score}. 
The point clouds in the quantitative testing configuration are subjected to a noise level of $2\%$. 
% As reported in Table \ref{SOTAResults}, SuperPC slightly outperforms the SOTA on the ShapeNet dataset but significantly surpasses it on the scene-level datasets. 
As shown in the second and third rows of \fref{fig:denoising}, the point cloud ground textures generated by SuperPC are noticeably less noisy compared to those produced by the SOTA point cloud denoising model - ScoreDenoise \cite{luo2021score}.

\vspace{-1pt}
\paragraph{Combination}
In the combination task, we integrate the SOTA models \cite{yu2023adapointr,he2023grad,luo2021score, nunes2024scaling, qu2024conditional, Mao_2024_CVPR} for each individual task in sequence to tackle the challenging combination task. 
As illustrated in \fref{fig:combination}, the combined approach with SOTA models falls short in achieving the final fully satisfactory map: despite successful upsampling, it leaves the rear window incomplete and introduces noise around the circled car. 
Conversely, SuperPC excels in producing more accurate point cloud maps, showcasing its enhanced capability to tackle complex combination tasks effectively.

\vspace{-1pt}
\paragraph{Colorization}
The point cloud colorization task cannot be evaluated alongside the other three tasks for two main reasons: first, its metrics differ significantly from the other tasks and cannot be adjusted to align with their metrics; second, combining the colorization task with the others would expand the point cloud data space from three dimensions to six, making it considerably more challenging for the diffusion model to learn the target data distribution. Therefore, we evaluate the colorization task separately using the same SuperPC model structure to demonstrate its effectiveness for this task. Details of the colorization task experiment can be found in the supplementary material Appendix \ref{sec:colorization}

\vspace{-3pt}
\subsection{Generalization Tests} \label{sec:Generalization_experiments}
\vspace{-5pt}

In addition to performance assessments, we evaluated the generalization capabilities of SuperPC and current SOTA models across various point cloud processing tasks. 
These evaluations were conducted through two critical experiments: object-to-scene and sim-to-real. 
These experiments are pivotal for practical applications of all the four tasks.

\paragraph{Object-to-Scene}
Traditionally, research has predominantly concentrated on object-level datasets such as ShapeNet \cite{chang2015shapenet}, often neglecting scene-level datasets. 
Yet, practical applications, such as indoor 3D reconstruction and urban digital twins, necessitate handling scene-scale point clouds. 
To address this, we trained SuperPC and various previous models on the object-level dataset (ShapeNet \cite{chang2015shapenet}) and subsequently fine-tuned and tested them on the scene-level dataset (TartanAir \cite{wang2020tartanair}). 
As shown in Table \ref{GenResults}, SuperPC demonstrates superior generalization from object-scale to scene-scale date compared to most state-of-the-art models, particularly in the combination task. 
% Only in the upsampling task and the object-to-scene setting does the SOTA model slightly outperform SuperPC.
% This discrepancy arises because SuperPC, as a multimodal framework, integrates both point cloud geometry and image semantic information. 
% Unlike point clouds, the structural and pattern differences between images of complex 3D scenes and simple objects are substantial, complicating the transfer of learned knowledge from object-scale to scene-scale settings. 
% Consequently, the image-modality part of the SuperPC model impedes its generalization ability in this object-to-scene challenge.

\paragraph{Sim-to-Real}
Another significant challenge is the sim-to-real gap. Learning-based models frequently perform well in simulated environments but struggle when applied to real-world scenarios. Our study is the first to train point cloud processing models on a simulated dataset (TartanAir \cite{wang2020tartanair}) and fine-tune and test them on a real-world dataset (KITTI-360 \cite{liao2022kitti}) to assess their capabilities in overcoming the sim-to-real gap. As illustrated in Table \ref{GenResults}, SuperPC outshines all SOTA models across all the point cloud processing tasks.

% \begin{figure}[t]
%     \centering
%     \includegraphics[width=0.8\linewidth]{images/generalization_test2.pdf}
%     \vspace{-9pt}
%     \captionsetup{width=.82\linewidth}
%     \caption{Plot of Generalization Experiment.}
%     \label{fig:generalization_test}
%     \vspace{-10pt}
% \end{figure}

\paragraph{Different Fine-tune Levels}
Furthermore, we fine-tuned both the integrated SOTA models and our SuperPC model using varying proportions of data from the TartanAir and KITTI-360 benchmarks to assess their generalization capabilities across both the object-to-scene and sim-to-real experiments. As depicted in \fref{fig:generalization_test}, increasing the fine-tuning levels enhances the performance of both SOTA models and SuperPC on the combination task. Notably, SuperPC consistently exhibits superior generalization ability across all levels of fine-tuning in both tracks. 
% Interestingly, in the sim-to-real experiment, the models converge more rapidly than in the object-to-scene experiment, suggesting that the gap between object-scale and scene-scale data may be harder to tackle than the sim-to-real gap for the point cloud processing tasks.

% %-------------------------------------------------------------------------
\vspace{-6pt}
\subsection{Additional Experiments} 
\vspace{-6pt}
In addition to the performance and generalization experiments, we present four more experiments in Appendix \ref{sec:additional_experiments} to further support our work. These include (1) the ablation study, (2) the complexity analysis, (3) the comparison between single-model and multi-model approaches using the same SuperPC network design, and (4) an evaluation of different integration orders when combining SOTA methods.

%-------------------------------------------------------------------------

\vspace{-6pt}
\section{Conclusion and Limitation}
\vspace{-6pt}
\paragraph{Conclusion}
We introduced SuperPC, a single diffusion model that excels in completion, upsampling, denoising, and colorization of point clouds. Leveraging the proposed TLC framework and SMF strategy, it outperforms specialized models in efficiency and effectiveness on complex scene-level datasets. SuperPC sets a new benchmark for generating high-quality point clouds, enhancing task interconnectivity, and paving the way for advanced 3D environmental analysis in future applications. 

\paragraph{Limitation}
Nevertheless, SuperPC’s efficiency can be further optimized by replacing the relatively heavy PointNet++ \cite{qi2017pointnet++} backbone with a sparse-tensor-based backbone. Furthermore, establishing a real-world object-level benchmark for this task would be highly valuable and convincing. In future work, researchers could explore the performance of a unified model across varying degrees of incompleteness derived from realistic scanning patterns.

\paragraph{Acknowledgments}
This work was supported by the DARPA award HR00112490426. Any opinions, findings, or conclusions expressed in this paper are those of the authors and do not necessarily reflect the views of DARPA.

{
    \small
    \bibliographystyle{ieeenat_fullname}
    \bibliography{main}
}

\clearpage
\setcounter{page}{1}

\appendix

% \maketitlesupplementary

% % % Some text content here to fill the space
% % \begin{figure*}[t]
% %     \centering
% %     \includegraphics[width=1\linewidth]{images/detailed_structure.pdf}
% %     \caption{The detailed structure of the SuperPC network.}
% %     \label{fig:main_model_structure}
% % \end{figure*}

\twocolumn[{%
\maketitlesupplementary
\renewcommand\twocolumn[1][]{#1}%
\vspace{-10pt}
\begin{center}
    \centering
    \captionsetup{type=figure}
    \includegraphics[width=1\textwidth]{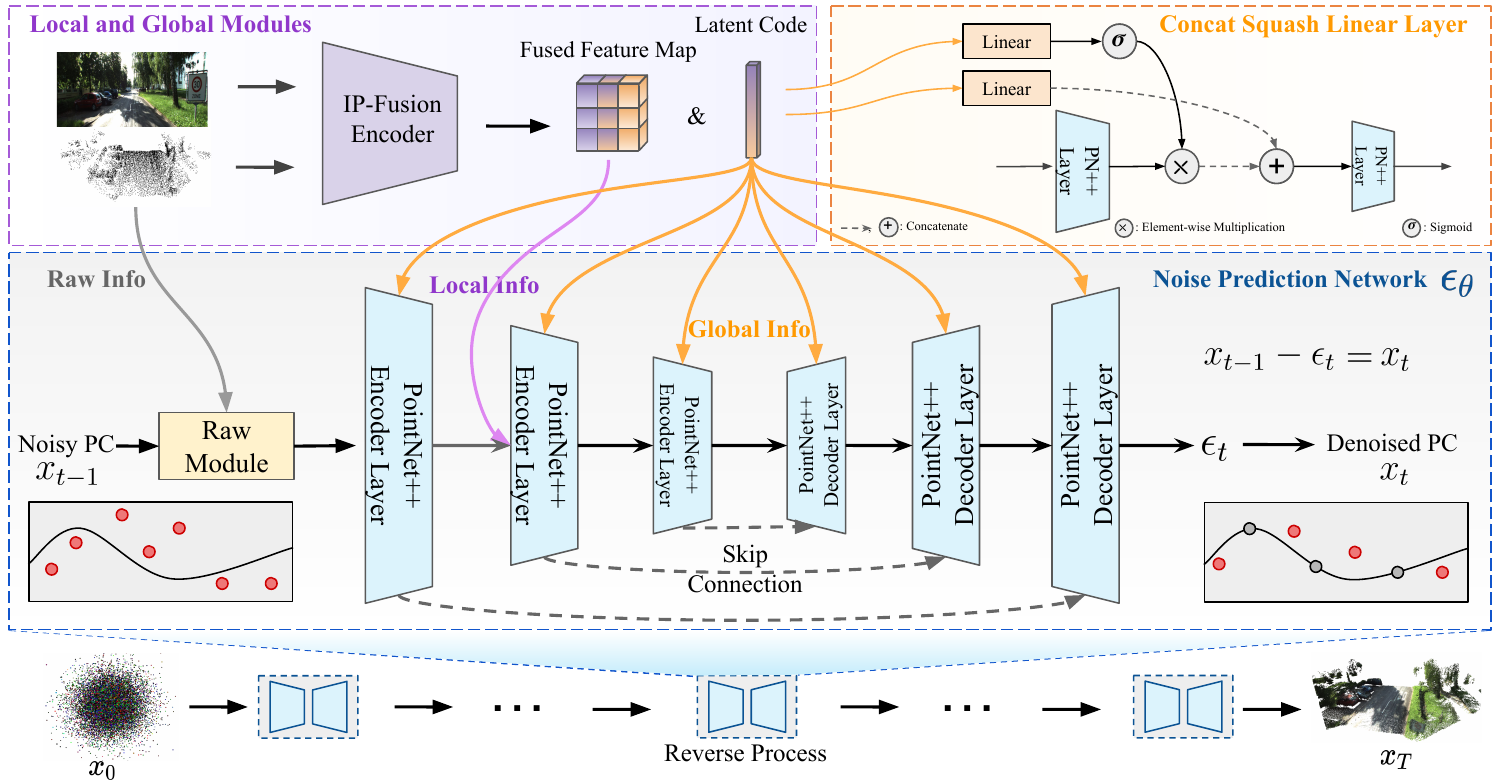}
    \vspace{-26pt}
    \captionof{figure}{The detailed structure of the SuperPC network.}
    \label{fig:main_model_structure}
\end{center}%
}]

% \twocolumn[{%
% % \maketitlesupplementary
% \renewcommand\twocolumn[1][]{#1}%
% % \vspace{-30pt}
% \begin{center}
%     \centering
%     \captionsetup{type=figure}
%     \includegraphics[width=1\textwidth]{images/detailed_structure.pdf} 
%     \captionof{figure}{The detailed structure of the SuperPC network.}
%     \label{fig:main_model_structure}
% \end{center}%
% }]

\section{Model Structure Details} \label{sec:detail_structure}

% The detailed structure of the SuperPC Model is illustrated in \fref{fig:main_model_structure}, where we specifically elaborate on the core diffusion noise prediction network and how information from two levels, obtained from the local and global modules, is incorporated into the core noise prediction network.
The SuperPC Model's detailed structure, including the core diffusion noise prediction network, is shown in \fref{fig:main_model_structure}. Moreover, it shows how information from raw, local, and global modules is integrated into the main network. We delve into the detailed explanation of them in this section.

For the core noise prediction network, we use the PointNet++ \cite{qi2017pointnet++} architecture as the backbone. However, we deviate from its original configuration, opting instead for the modified version introduced in PDR \cite{lyu2021conditional} and DDPMPU \cite{qu2024conditional}. This adaptation is due to the original PointNet++ network's inability to effectively process point features from clouds resembling Gaussian noise. The network comprises a three-level PointNet++ encoder and decoder structure. In the encoding stages, we set the number of neighbors, K, as 16 for set abstraction purposes. In the decoding stages, K is set to 8 to facilitate feature propagation.

Since we have already detailed how raw information is incorporated into the core network in \sref{sec:raw_module}, we will next explain how information from local and global levels is integrated into the core network. 
After passing the input image and point cloud into the local and global modules, we get the local feature map and the global latent code. 
The local feature map $M(N_l, 3+C_l)$ can essentially be viewed as $N_l$ points each possessing $3+C_l$ features, which shares the same format as the output $P_1(N_1, 3+C_1)$ of the core network's first layer. 
We employ the point spatial interpolation method mentioned in \sref{sec:raw_module} to align the local information from the local feature map with the output of the first layer of the core network.
The global latent code $\boldsymbol{z}(1,1024)$ is added to the core network via the concatsquash layer \cite{grathwohl2018ffjord} suggested by Luo \cite{luo2021diffusion}, which is defined as:
\begin{equation}
    P_1' = CS(P_1, t, \boldsymbol{z}) = P_1 \odot \sigma(\boldsymbol{W_1c} + \boldsymbol{b_1}) + \boldsymbol{W_2c},
\end{equation}
where \( P_1 \) represents the input to the layer, while \( P_1' \) signifies the output. Here, \( \boldsymbol{c} = [t, \sin(t), \cos(t), \boldsymbol{z}] \) constitutes the context vector including the embedding of time $t$ and the global latent code $\boldsymbol{z}$, with \( \sigma \) indicating the sigmoid function. The parameters \( \boldsymbol{W_1}, \boldsymbol{W_2} \) and \( \boldsymbol{b_1} \) are all subject to optimization during training.
The code with detailed implementation will be released upon the paper's acceptance.

% \section{Implementation Details} \label{implementation}
\section{Benchmark Details} \label{sec:dataset_details}
\subsection{ShapeNet}
For the object-level benchmark, we render the images of the 3D objects in the ShapeNet \cite{chang2015shapenet} following the setting and method provided by \cite{xu2019disn, choy20163d} since SuperPC integrates the image information. 
In addition to the rendered images, we sample the 3D models from the ShapeNet Core Dataset to create ground truth point clouds that are paired with the images. 
We appreciate the perspective of the ShapeNet55/34 dataset \cite{yu2023adapointr} on the necessity of assessing model generalization performance across unseen categories. 
We chose thirteen categories used by \cite{xu2019disn, choy20163d} from the ShapeNet Core dataset. Ten of these categories are designated as `seen' for training and validation, with the remaining three categories earmarked as `unseen' for testing, to evaluate the models' ability to generalize.
Each ground truth point cloud is standardized to contain 8,192 points, following the specifications stated in the ShapeNet55/34 dataset \cite{yu2023adapointr}.

\subsection{TartanAir}
% Existing point cloud processing datasets typically concentrate on simple, synthetic objects, which limits their applicability to real-world scenarios involving complex, multi-object scenes.
Existing point cloud processing datasets typically concentrate on simple, synthetic objects, which are insufficient to verify the performance of models in complex scenarios.
To bridge this gap and demonstrate the effectiveness of our work, we introduce a scene-level benchmark utilizing the TartanAir dataset \cite{wang2020tartanair}, aimed at evaluating model performance in complex environments. 
It features fifteen diverse indoor and outdoor environments, covering different seasons and lighting conditions, and is derived from 176 sequences totaling over 600,000 frames. 
This results in a comprehensive dataset of 85,618 point clouds paired with images, providing a robust benchmark for assessing point cloud processing tasks.
Most existing point cloud processing datasets focus solely on simple, virtual objects. However, in real-world applications, the point clouds we often need to process consist of complex scenes with many objects. Therefore, we propose constructing a scene-level benchmark based on the TartanAir \cite{wang2020tartanair} dataset to evaluate the effectiveness of models and methods in performing point cloud processing tasks within complex scenarios.
% In our experiments, we utilize fifteen diverse environments, encompassing both indoor and outdoor settings across various seasons and lighting conditions. Drawing from 176 distinct sequences that comprise over 600,000 frames, our benchmarks incorporate 85,618 point clouds accompanied by images. 
TartanAir provides the RGB and Depth images in eighteen photo-realistic simulation environments. 
We generate the raw point clouds based on the RGBD images with a depth-limit truncation to remove those points with huge depth values like the points representing the sky. 
These raw point clouds are downsampled to 46080 points to serve as the ground truth, accommodating the memory constraints of training baseline models \cite{yu2023adapointr, luo2021score} and meeting the requirements for Earth Mover's Distance (EMD) calculations. 
% The distribution of environments for training, validation, and testing follows a ratio of $9:3:3$.

\subsection{KITTI-360} \label{sec:kitti360_benchmark}
Although the TartanAir Benchmark provides data for evaluating scene-level performance, it is based on simulations. 
To better validate the effectiveness of our method in real-world scenarios, we also include a real-world, scene-level point cloud processing benchmark based on KITTI-360 \cite{liao2022kitti}.
It provides high-quality images and accurate accumulated point clouds. 
We stitch together the accumulated point clouds from each sequence to create dense global maps and, based on the pose information, crop out dense local point clouds from these maps. 
Each local point cloud is then downsampled to 46,080 points to serve as ground truth and matched with the corresponding frame's image to form the KITTI-360 benchmark dataset used in this work.

\section{Metrics Details} \label{sec:metrics_details}
\subsection{Density-aware Chamfer Distance} \label{sec:DCD}

DCD \cite{wu2021density} improves the evaluation of visual quality for 3D shape generation tasks by considering the density of points in a point cloud, unlike the traditional Chamfer Distance. 
DCD's formulation takes into account both the point-to-point distances and the point densities, providing a more discriminative measure.
The DCD between two point clouds \( S_1 \) and \( S_2 \) is given by the following equation:
\begin{equation}
    \resizebox{0.88\linewidth}{!}{
        $\displaystyle
        \begin{aligned}
            d_{DCD}(S_1, S_2) = \frac{1}{2} \Bigg( & \frac{1}{|S_1|} \sum_{x \in S_1} \min_{y \in S_2} \left( 1 - e^{-\alpha||x-y||^2} \right) \\
            + & \frac{1}{|S_2|} \sum_{y \in S_2} \min_{x \in S_1} \left( 1 - e^{-\alpha||x-y||^2} \right) \Bigg)
        \end{aligned}
        $
    }
\end{equation}

In this formula, \( S_1 \) and \( S_2 \) are the two sets of points that represent point clouds. 
The cardinalities \( |S_1| \) and \( |S_2| \) indicate the number of points in each set. The variables \( x \) and \( y \) correspond to the points in \( S_1 \) and \( S_2 \), respectively. 
The term \( ||x-y||^2 \) is the squared Euclidean distance between the points \( x \) and \( y \). 
The exponential term \( e^{-\alpha||x-y||^2} \) is used to calculate a distance that is sensitive to the point density, with \( \alpha \) acting as a temperature scalar that influences the sensitivity of the distance to point density variations. 
The minimum function \( \min \) finds the nearest neighbor distance, ensuring that each point in one set is compared to its closest point in the other set. 
This formulation indicates that the DCD is not just the average nearest neighbor distance but also incorporates a normalization based on the local density of points, which helps to prevent the measure from being too sensitive to outliers and provides a better representation of the actual shape and structure of the point clouds.

\section{Additional Experiments} \label{sec:additional_experiments}
In the main text, we have already discussed the performance and generalization experiments. To provide a more comprehensive evaluation, we will provide (1) the ablation study in Appendix \ref{sec:ablation}, (2) the complexity analysis in Appendix \ref{sec:complexity_analysis}, (3) the comparison between the single unified model and different combinations of multiple models for individual tasks with the same SuperPC framework in Appendix \ref{sec:single_vs_multi}, (4) an evaluation of different integration orders when combining SOTA methods in Appendix \ref{sec:different_integration}, and (5) the colorization task results in Appendix \ref{sec:colorization}.

\setlength\heavyrulewidth{0.32ex}
\begin{table*}[t]
\caption{Ablation Study of the Image-Point Fusion and Three-Level-Info conditions.} 
\label{AblationResults}
\centering
 \resizebox{2.1\columnwidth}{!}{
\begin{tabular}{ccccccccccccccc}
\toprule
\multicolumn{2}{c}{Fusion Stage} & \multicolumn{3}{c}{Condition Module} & \multicolumn{3}{c}{ShapeNet \cite{chang2015shapenet}} & \multicolumn{3}{c}{TartanAir \cite{wang2020tartanair}} & \multicolumn{3}{c}{KITTI-360 \cite{liao2022kitti}} \\ \cmidrule(lr){0-1} \cmidrule(lr){3-5} \cmidrule(lr){6-8} \cmidrule(lr){9-11} \cmidrule(lr){12-14}
Early       &Deep          &Raw        &Local       &Global       &DCD($\downarrow$)&EMD($\downarrow$)&F1($\uparrow$) &DCD($\downarrow$)&EMD($\downarrow$)&F1($\uparrow$) &DCD($\downarrow$)&EMD($\downarrow$)&F1($\uparrow$) \\ \midrule

$\crossmark$  &$\checkmark$ &$\checkmark$  &$\checkmark$   &$\checkmark$      &0.661            &8.59            &0.254            &0.822            &11.73            &0.183            &0.935            &20.89            &0.177         \\
$\checkmark$  &$\crossmark$ &$\checkmark$  &$\checkmark$   &$\checkmark$      &0.623            &8.24            &0.295            &0.796            &11.45            &0.239            &0.896            &20.56            &0.203         \\
$\checkmark$  &$\checkmark$ &$\crossmark$  &$\checkmark$   &$\checkmark$      &0.648            &8.31            &0.263            &0.811            &12.68            &0.201            &0.925            &23.79            &0.181         \\
$\checkmark$  &$\checkmark$ &$\checkmark$  &$\crossmark$   &$\checkmark$      &0.594            &7.79            &0.375            &0.658            &9.68             &0.319            &0.794            &17.28            &0.227         \\
$\checkmark$  &$\checkmark$ &$\checkmark$  &$\checkmark$   &$\crossmark$      &0.693            &8.96            &0.248            &0.697            &10.43            &0.298            &0.852            &19.71            &0.205         \\ 
$\checkmark$  &$\checkmark$ &$\checkmark$  &$\checkmark$   &$\checkmark$      &\textbf{0.476}   &\textbf{2.21}   &\textbf{0.409}   &\textbf{0.558}   &\textbf{3.527}   &\textbf{0.384}   &\textbf{0.681}   &\textbf{9.58}    &\textbf{0.365}\\ \bottomrule
% \vspace{-38pt}
\end{tabular}
}
\end{table*}

\setlength\heavyrulewidth{0.36ex}
\begin{table*}[t]
\caption{Comparison between the single unified model and different combinations of multiple models for individual tasks (all the models use the same SuperPC framework for fairness).}
\label{Single_vs_Multi}
\centering
% ShapeNet
\scalebox{0.94}{
\begin{tabular}{ccccccccccc}
\toprule 
\multirow{3}{*}{Different Combinations} & \multicolumn{3}{c}{\textbf{ShapeNet} \cite{chang2015shapenet}} & \multicolumn{3}{c}{\textbf{TartanAir} \cite{wang2020tartanair}} & \multicolumn{3}{c}{\textbf{KITTI-360} \cite{liao2022kitti}} \\ \cmidrule(lr){2-4} \cmidrule(lr){5-7} \cmidrule(lr){8-10}
                    & DCD ($\downarrow$)         & EMD ($\downarrow$)       & F1 ($\uparrow$)       & DCD ($\downarrow$)          & EMD ($\downarrow$)       & F1 ($\uparrow$)       & DCD ($\downarrow$)       & EMD ($\downarrow$)       & F1 ($\uparrow$)       \\ \midrule
SPC(U) + SPC(C) + SPC(D)  & 0.506            & 2.46           & 0.374            & 0.574             & 3.67             & 0.354             & 0.715             & 11.92          & 0.335            \\

SPC(C) + SPC(D+U)         & 0.492            & 2.37           & 0.388            & \underline{0.561} & \underline{3.59} & \underline{0.373} & \underline{0.692}             & \underline{10.17}          & \underline{0.352}            \\

SPC(D) + SPC(C+U)         & 0.495            & 2.41           & 0.383            & 0.571             & 3.63             & 0.364             & 0.707             & 10.84          & 0.343        \\ 

SPC(U) + SPC(C+D)         &\underline{0.489} &\underline{2.32}&\underline{0.391} & 0.564             & 3.61             & 0.368             & 0.698             & 10.53          & 0.347            \\ 

SPC(C+U+D)                & \textbf{0.476}   & \textbf{2.21}  & \textbf{0.409}   & \textbf{0.558}    & \textbf{3.53}    & \textbf{0.384}    & \textbf{0.681}    & \textbf{9.58}  & \textbf{0.365}     \\ \bottomrule
\end{tabular}
}
% \vspace{-6pt}
\end{table*}

\subsection{Ablation Study} \label{sec:ablation}
The ablation studies are performed to evaluate the effectiveness of the five critical components in our model: the dual-spatial early fusion, the attention-based deep fusion, the raw module, the local module, and the global module. 

\paragraph{Early Fusion and Deep Fusion}
% To demonstrate the importance of our spatial-mixed-fusion strategy, we eliminated the image modality at the early-fusion stage and deep-fusion stage separately, such as the image feature projection, the image encoder, and so on, as discussed in \sref{sec:raw_module} and \sref{sec:local_module}. As shown in \tref{AblationResults}, removing any fusion stage leads to a noticeable drop in the overall performance across all three datasets.
To demonstrate the importance of our spatial-mixed-fusion strategy, we conducted an ablation study by removing the image modality at two critical fusion stages: the early-fusion stage (image feature projection) and the deep-fusion stage (image encoder with the cross-attention module), as described in \sref{sec:raw_module} and \sref{sec:local_module}. 
As shown in \tref{AblationResults}, excluding either fusion stage results in a significant decline in overall performance across all three benchmarks, underscoring the importance of incorporating both the early fusion and the deep fusion.

\paragraph{Raw, local, and global module}
The evaluation of the three-level modules involves removing each of these components individually.
Excluding any of these elements disrupts the integrity of the three-level-conditioned framework, leading to a marked deterioration in overall performance as shown in \tref{AblationResults}. This effect is most pronounced with the raw module, as its exclusion leads to a notable decline in performance.
% Generally speaking, every module plays a significant role in building the three-level-conditioned, two-modal fusion framework, significantly enhancing the robustness and general performance of our model.
Generally speaking, every module plays a significant role in building the TLC framework and GMF strategy.

% \setlength\heavyrulewidth{0.36ex}
% \begin{table}[h]
% \aboverulesep=0ex % Solution part 1 of 3
% \belowrulesep=0ex % Solution part 1 of 3
% \caption{Complexity of SOTAs combination and SuperPC with different reverse steps. PU, PC, and PD are the SOTAs of upsampling \cite{he2023grad}, completion \cite{yu2023adapointr}, and denosing \cite{luo2021score} on ShapeNet. All the results were tested on an NVIDIA GeForce RTX 3090 GPU.} 
% % \vspace{-6pt}
% \label{complexity}
% \centering
% \scalebox{0.7}{
% \begin{tabular} {l|ccc|ccc}
% \toprule
% {Method}             & {Params}        & {FLOPs}          & {$t_{inf}$}     & {DCD($\downarrow$)}& {EMD($\downarrow$)}& {F1($\uparrow$)}  \\ \midrule
% PU+PC+PD             & \textbf{33.36 M}& 593.6 G          & 3.92 s          & 0.521              & 2.93               & 0.362        \\\midrule
% SuperPC (50 steps)  & 36.78 M         & \textbf{93.5 G}  & \textbf{0.76 s} & 0.507              & 2.64               & 0.379        \\ 
% SuperPC (100 steps) & 36.78 M         & 183.4 G          & 1.38 s          & 0.492              & 2.43               & 0.388        \\ 
% SuperPC (1000 steps)& 36.78 M         & 1809.6 G         & 14.69 s         & \textbf{0.476 }    & \textbf{2.21}      & \textbf{0.409}\\\bottomrule
% \end{tabular}
% }
% \vspace{-6pt}
% \end{table}

\setlength\heavyrulewidth{0.36ex}
\begin{table}[t]
\aboverulesep=0ex % Solution part 1 of 3
\belowrulesep=0ex % Solution part 1 of 3
\caption{Complexity of SOTAs combination and SuperPC with different reverse steps. PU, PC, and PD are the SOTAs of denosing \cite{luo2021score} , completion \cite{yu2023adapointr}, and upsampling \cite{qu2024conditional} on ShapeNet. All the results were tested on an NVIDIA GeForce RTX 3090 GPU.} 
\label{complexity}
\centering
\scalebox{0.83}{
\begin{tabular} {l|ccc|c}
\toprule
{Method}             & {Params}        & {FLOPs}          & {$t_{inf}$}     & {DCD($\downarrow$)} \\ \midrule
PD+PC+PU             & \textbf{33.36 M}& 593.6 G          & 3.92 s          & 0.462         \\\midrule
SuperPC (50 steps)  & 36.78 M         & \textbf{93.5 G}  & \textbf{0.76 s} & 0.441         \\ 
SuperPC (100 steps) & 36.78 M         & 183.4 G          & 1.38 s          & 0.412          \\ 
SuperPC (1000 steps)& 36.78 M         & 1809.6 G         & 14.69 s         & \textbf{0.387 }  \\\bottomrule
\end{tabular}
}
\vspace{-8pt}
\end{table}

\subsection{Complexity Analysis} \label{sec:complexity_analysis}
To fulfill the goal of the combination task, previous single-task models \cite{he2023grad, qu2024conditional, yu2023adapointr, luo2021score} could only be sequentially interpreted together to accomplish point cloud upsampling, completion, and denoising step by step. In contrast, our SuperPC is capable of completing the entire combination task within one single model. Therefore, theoretically, not only can it achieve higher performance as proven in Section 3.1, but it also requires less computational consumption and shorter inference time. As shown in \tref{complexity}, SuperPC demonstrates higher performance across all three metrics, along with lower FLOPs and inference time ($t_{inf}$) compared with the combination of the SOTAs \cite{luo2021score,yu2023adapointr,qu2024conditional} of the three single tasks, whether setting the reverse steps of the diffusion model to 50 or 100. 
Due to the principles of diffusion models, more reverse steps can improve the quality of inference but also require more computation and inference time. 
In practical applications, using 100 steps allows the model to generate high-quality point clouds within a relatively short inference time. 
% To be honest, the current model still falls short in efficiency, particularly in terms of parameter count and during a large number of inference steps. 
Moving forward, we aim to further enhance the model's efficiency by either refining the point diffusion mechanism or replacing the current complex point cloud learning backbone \cite{qi2017pointnet++} with the sparse-tensor-based backbone like Minkowski Engine \cite{choy20194d}.

\setlength\heavyrulewidth{0.36ex}
\begin{table*}[t]
\caption{Results of different SOTAs integration methods on the combination task.}
\vspace{-10pt}
\label{DiffIntegrateResults}
\centering
\scalebox{0.92}{
\begin{tabular}{cccccccccccc}
\toprule
\multirow{3}{*}{Task} & \multirow{3}{*}{Methods} & \multicolumn{3}{c}{ShapeNet \cite{chang2015shapenet}} & \multicolumn{3}{c}{TartanAir \cite{wang2020tartanair}} & \multicolumn{3}{c}{KITTI-360 \cite{liao2022kitti}} \\ \cmidrule(lr){3-5} \cmidrule(lr){6-8} \cmidrule(lr){9-11}  
                      &                         & DCD ($\downarrow$)         & EMD ($\downarrow$)       & F1 ($\uparrow$)       & DCD ($\downarrow$)          & EMD ($\downarrow$)       & F1 ($\uparrow$)       & DCD ($\downarrow$)       & EMD ($\downarrow$)       & F1 ($\uparrow$)      \\ \midrule
\multirow{4}{*}{Combination} 
& PD$\rightarrow$PC$\rightarrow$PU                 &\underline{0.489}& 2.64           &\underline{0.391}& 0.612           & 3.93            & 0.125           & 0.749           & 10.18            & 0.254            \\ 
& PU$\rightarrow$PD$\rightarrow$PC                 & 0.497           &\underline{2.36}& 0.375           & 0.609           & 3.82            & 0.130           & 0.763           & 10.29           & 0.248           \\ 
& PU$\rightarrow$PC$\rightarrow$PD                 & 0.521           & 2.93           & 0.362           &\underline{0.583}&\underline{3.64} &\underline{0.139}&\underline{0.725}&\underline{10.06}&\underline{0.266}\\ 
& SuperPC (ours)                                  &\textbf{0.476}   &\textbf{2.21}   &\textbf{0.409}   &\textbf{0.558}   &\textbf{3.527}   &\textbf{0.154}   &\textbf{0.681}&\textbf{9.58}&\textbf{0.287}\\ \bottomrule
\vspace{-18pt}
\end{tabular}
}
\end{table*}

% \begin{table*}[t]
%     \centering
%     \caption{Colorization Experiment.}
%     \begin{tabular}{ccccc}
%         \toprule
%         \multirow{3}{*}{\textbf{Methods} [32]} & \multicolumn{2}{c}{ShapeNet} & \multicolumn{2}{c}{TartanAir} & \multicolumn{2}{c}{KITTI-360} \\ \cmidrule(lr){2} \cmidrule(lr){3} \cmidrule(lr){4}
%                               & MSE ($\downarrow$)              & MSE ($\downarrow$)               & MSE ($\downarrow$)       \\ \midrule 

%         Learning-based [33]   
%                         & 0.0316          & 0.0536          & 0.0536          \\ 
%         Geometry-based   
%                         & 0.0433          & 0.0142          & 0.0142         \\ 
%         SuperPC         &\textbf{0.0158}  &\textbf{0.0129}  &\textbf{0.0129}   \\ \bottomrule
%     \end{tabular}
%     \label{tab:colorization_experiment}
% \end{table*}

% --------------- Combination ----------------
\begin{figure*}[!t]
    \centering
    \vspace{-6pt}
    \includegraphics[width=1\linewidth]{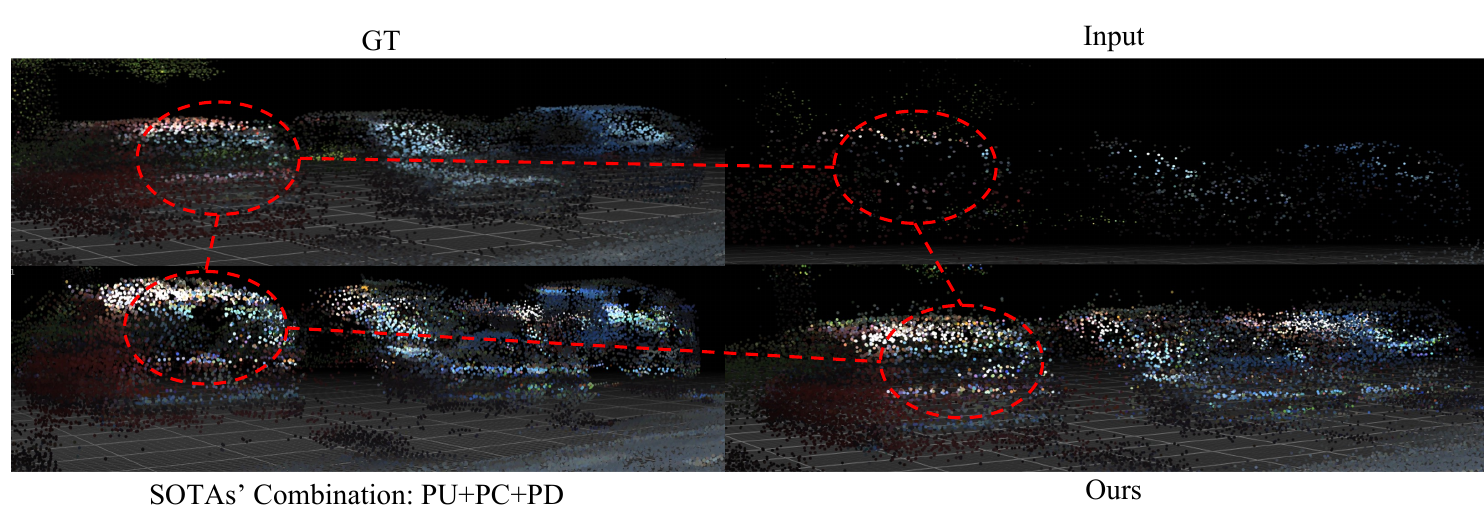} 
    \vspace{-23pt}
    \caption{The quality results of SuperPC and SOTAs on the combination task.}
    \label{fig:combination}
    \vspace{-10pt}
\end{figure*}
% --------------- Combination ----------------

\subsection{Single model vs multiple-models-combination} \label{sec:single_vs_multi}
\vspace{-5pt}
Despite significant advancements, prior approaches \cite{luo2021score, he2023grad, yu2023adapointr, liu2019pccn} predominantly tackle each of these tasks—completion, upsampling, denoising, and colorization—independently. However, such isolated strategies overlook the inherent interdependence among defects including incompleteness, low resolution, noise, and lack of color, which frequently coexist and influence one another.

Currently, there is no single model capable of addressing all four tasks simultaneously. A unified model offers not only computational efficiency but also the ability to prevent error accumulation across tasks while leveraging their interconnectivity to mutually enhance performance. For instance, as illustrated in  \fref{fig:combination}, errors from a completion model often propagate to subsequent upsampling.
In addition to qualitative observations highlighting the limitations of combining multiple specialized models, we conducted extensive quantitative experiments to substantiate this claim. Specifically, we compared our single unified model with various combinations of multiple models for individual tasks, ensuring a fair comparison by implementing all models within the same SuperPC framework. As demonstrated in \tref{Single_vs_Multi}, the single unified model consistently outperforms all combinations across three benchmarks.
These findings underscore the necessity of a single, integrated model capable of simultaneously addressing all four tasks.

\vspace{-3pt}
\subsection{Different SOTAs integration methods} \label{sec:different_integration}
\vspace{-5pt}
In the combination task, we integrate the SOTA models \cite{yu2023adapointr, he2023grad, luo2021score} for each individual task in various reasonable sequences, as shown in Table \ref{DiffIntegrateResults}. The sequence starting with upsampling (PU) \cite{he2023grad}, followed by completion (PC) \cite{nunes2024scaling}, and ending with denoising (PD) \cite{luo2021score} yields relatively better outcomes compared to other combinations in the scene-scale datasets - TartanAir and KITTI-360. However, the sequence of "PD$\rightarrow$PC$\rightarrow$PU" shows better performance on the object-level dataset - ShapeNet. Obviously, SuperPC surpasses all the integration methods across the three datasets.

\begin{table}[t]
% \vspace{-3pt}
    \centering
    \caption{Colorization Experiment. MSE is used as the metric.}
    \vspace{-5pt}
    \scalebox{0.92}{
    \begin{tabular}{ccccc}
        \toprule
        \multirow{1}{*}{\textbf{Methods}} & \multicolumn{1}{c}{ShapeNet} & \multicolumn{1}{c}{TartanAir} & \multicolumn{1}{c}{KITTI-360} \\      \midrule 

        Learning-based \cite{liu2019pccn}   
                        & 0.0316          & 0.0429          & 0.0536          \\ 
        Geometry-based   
                        & 0.0276          & 0.0131          & 0.0142         \\ 
        SuperPC         &\textbf{0.0102}  &\textbf{0.0117}  &\textbf{0.0129}   \\ \bottomrule
    \end{tabular}
    }
    \vspace{-11pt}
    \label{tab:colorization_experiment}
\end{table}
\vspace{-3pt}
\subsection{Colorization Task} \label{sec:colorization}
\begin{figure}[h]
    \centering
    \includegraphics[width=1\linewidth]{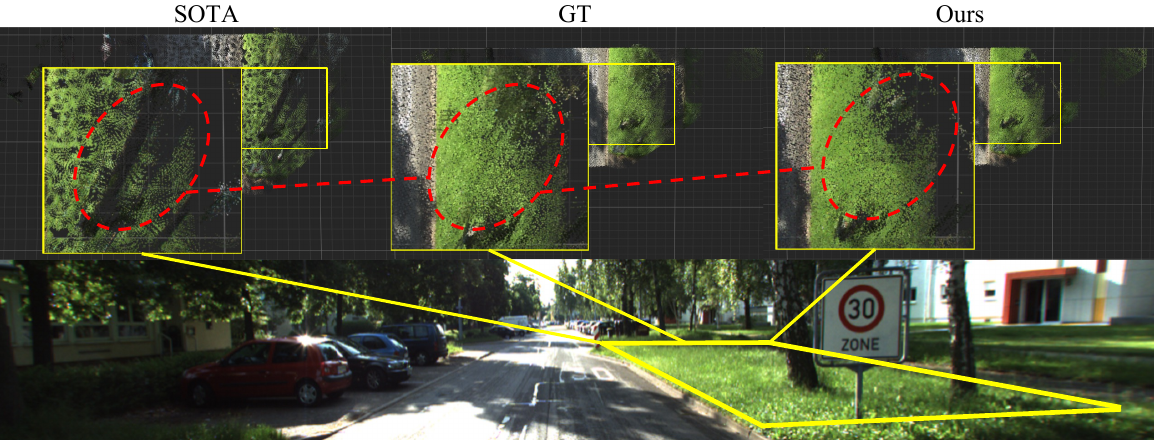} 
    \caption{The quality results of SuperPC and SOTA (projection) on the point cloud colorization task with zoom in details on the generated green color of the grass field.}
    \label{fig:colorization2}
    % \vspace{5pt}
\end{figure}

\begin{figure}[h]
    \centering
    \includegraphics[width=1\linewidth]{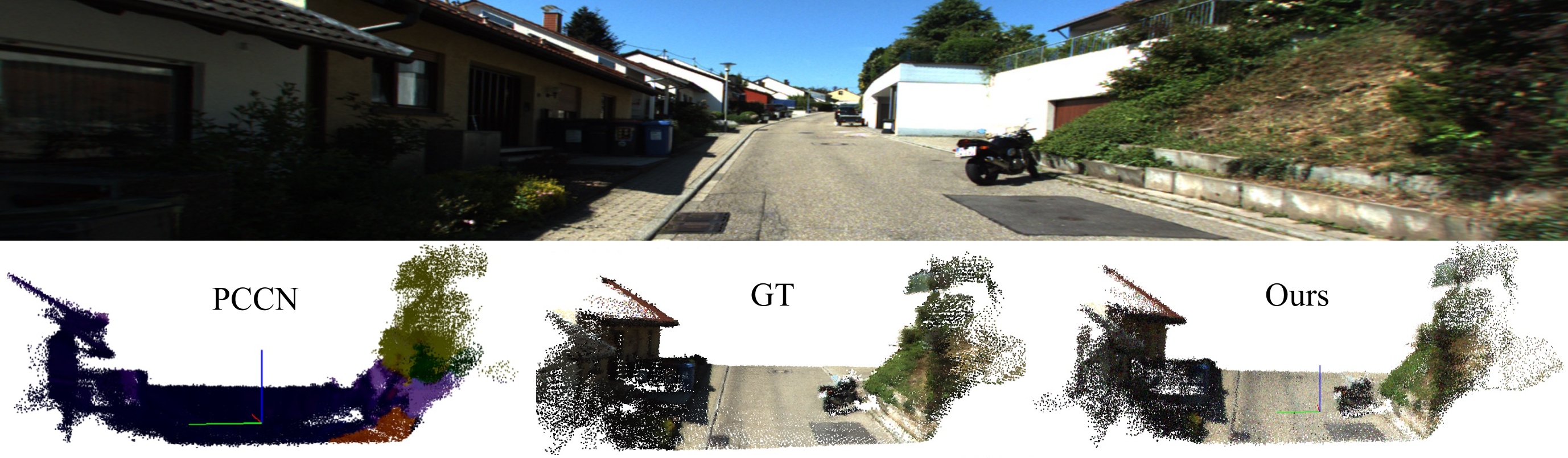} 
    \caption{The quality results of SuperPC method and baseline learning method (PCCN \cite{liu2019pccn}).}
    \label{fig:colorization}
\end{figure}

\vspace{-3pt}
In the colorization task, we evaluate the qualitative performance of the SuperPC compared to the baseline model \cite{liu2019pccn} and the SOTA - geometry-based method.
The learning-based baseline model sometimes generates weird unreal colors as shown at the left bottom of \fref{fig:colorization}.
The projection-based method exhibits limitations in rendering colors for obscured scenes. A specific instance highlighted in \fref{fig:colorization2} reveals its failure to accurately colorize grass hidden by a tree.
In contrast, SuperPC effectively predicts the colors for occluded areas, producing the point cloud that closely aligns with the ground truth texture and colors. Additionally, as shown in \tref{tab:colorization_experiment}, SuperPC outperforms both the learning-based and the geometry-based methods.
% Overall, SuperPC demonstrates superior robustness in the point cloud colorization task.

% \clearpage
\subsection{Experiment on Observation Incompleteness}
\label{sec:varing_incomplete}

We generate three distinct levels of observation incompleteness by stitching point clouds from one, three, and five adjacent frames, followed by cropping them to maintain consistency in camera pose and field of view. 
The PC completion performance of SuperPC is compared against the current SOTA method using the average results across these three levels of incompleteness, evaluated on two scene-level datasets: TartanAir \cite{wang2020tartanair} and KITTI-360 \cite{liao2022kitti}.

As shown in Table~\ref{tab:varing_incomplete_evaluation}, SuperPC consistently outperforms the SOTA method across both evaluated datasets, demonstrating superior robustness and effectiveness in handling varying degrees of observation incompleteness.
\vspace{-5pt}
\begin{table}[h]
\centering
\small % Reduce the font size of the table
\setlength{\tabcolsep}{4pt} % Adjust column separation for a more compact table
\renewcommand{\arraystretch}{0.9} % Adjust row height for a more compact table
\begin{tabular}{|c|c|c|c|c|c|}
\hline
\textbf{Datasets} & \textbf{Methods} & \textbf{CD} & \textbf{DCD} & \textbf{EMD} & \textbf{F1} \\ 
\hline
\multirow{2}{*}{KITTI-360 \cite{liao2022kitti}} & LiDiff \cite{nunes2024scaling} & 9.41 & 0.693 & 9.82 & 0.247 \\ \cline{2-6}
& SuperPC & \textbf{8.63} & \textbf{0.667} & \textbf{9.24} & \textbf{0.298} \\ 
\hline
\multirow{2}{*}{TartanAir \cite{wang2020tartanair}} & LiDiff \cite{nunes2024scaling} & 7.91 & 0.631 & 4.52 & 0.296 \\ \cline{2-6}
& SuperPC & \textbf{7.04} & \textbf{0.597} & \textbf{4.15} & \textbf{0.327} \\
\hline
\end{tabular}
\vspace{-8pt}
\caption{Incomplete observations evaluation.}
\vspace{-12pt}
\label{tab:varing_incomplete_evaluation}
\end{table}
\vspace{-5pt}

\subsection{Combination Experiment on PCN Dataset}
\label{sec:varing_incomplete}

We present a brief performance comparison between SuperPC and SOTA methods combination on the PCN dataset \cite{yuan2018pcn}. As shown in \tref{tab:PCN_dataset} below, the results demonstrate that SuperPC significantly outperforms SOTA methods on the combination task, which is our main contribution.

\vspace{-5pt}
\begin{table}[h]
\centering
\small
\setlength{\tabcolsep}{4pt} % Adjust column spacing
\renewcommand{\arraystretch}{0.9} % Adjust row height
\begin{tabular}{|c|c|c|c|c|}
\hline
\textbf{Methods} & \textbf{CD} & \textbf{DCD} & \textbf{EMD} & \textbf{F1} \\ \hline
\cite{he2023grad}$\rightarrow$\cite{nunes2024scaling}$\rightarrow$\cite{luo2021score}       & 11.03           & 0.495          & 3.44          & 0.592 \\ \hline
    SuperPC              & \textbf{10.12}  & \textbf{0.432} & \textbf{2.13}  & \textbf{0.675}    \\ \hline
\end{tabular}
\vspace{-8pt}
\caption{Combination task performance on PCN dataset.}
\vspace{-8pt} % Adjust spacing below the table
\label{tab:PCN_dataset}
\end{table}
\vspace{-5pt}

% --------------- More Qualitative Samples ----------------
\clearpage
\onecolumn
\section{More Qualitative Samples}\label{sec:more_quantitative}

% --------------- Colorization ----------------
\vspace{38pt}

\vspace{38pt}
% --------------- Colorization ----------------

\begin{figure}[h]
    \centering
    \includegraphics[width=0.96\linewidth]{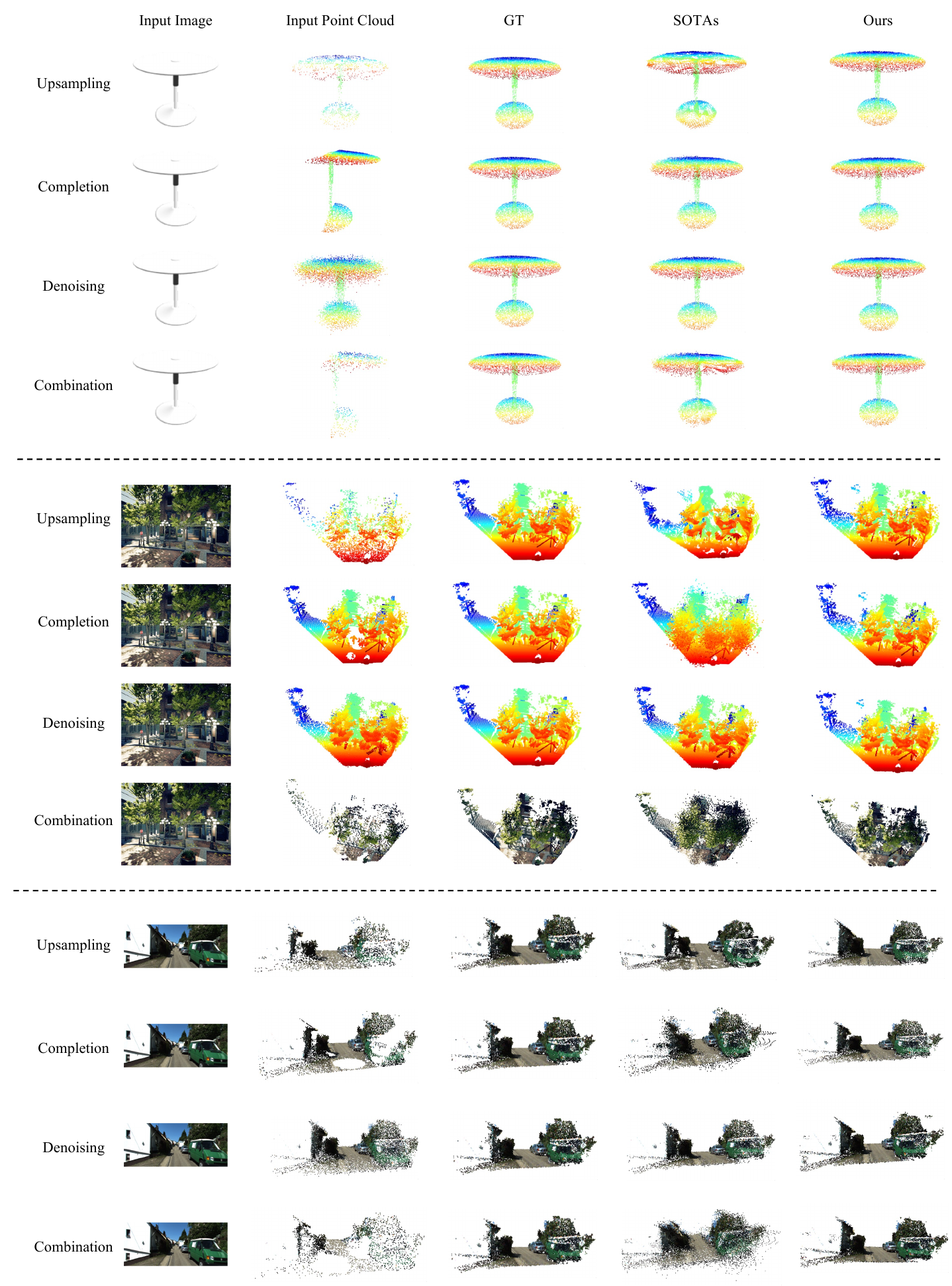} 
    \vspace{-10pt}
    \caption{More qualitative results on the ShapeNet, TartanAir, and KITTI-360 dataset.}
    % \vspace{-46pt}
    \label{fig:appendix}
\end{figure}

\begin{figure}[h]
    \centering
    \includegraphics[width=1\linewidth]{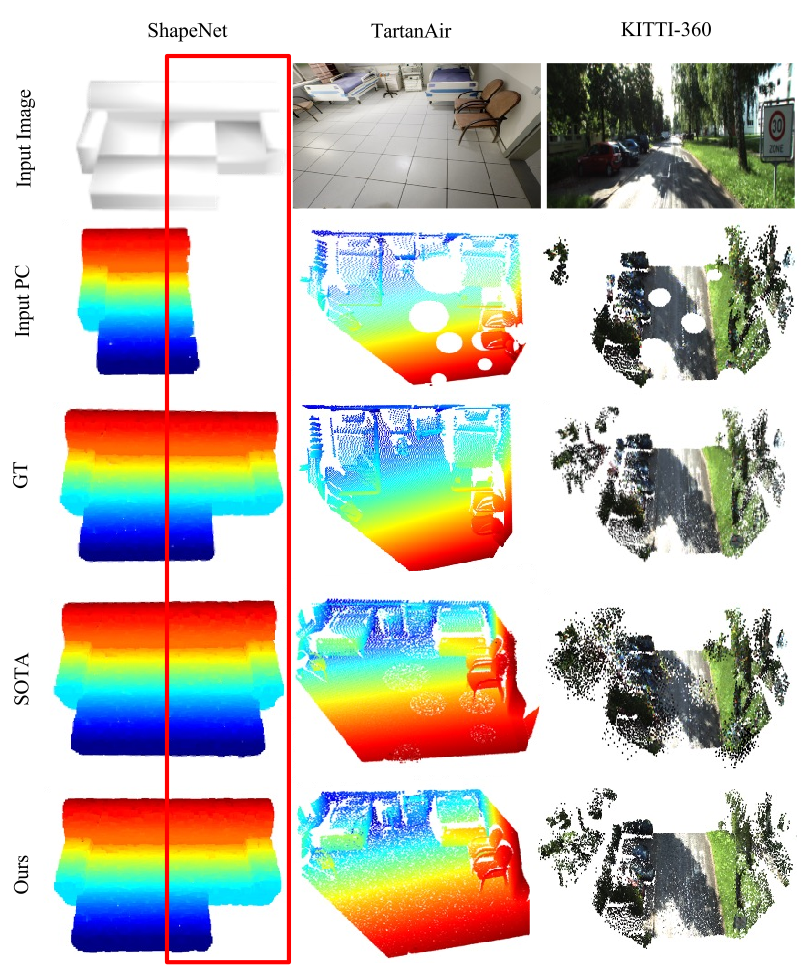} 
    \caption{Zoom-in figure of the completion task qualitative results.}
    \vspace{-46pt}
    \label{fig:completion_appendix}
\end{figure}

\begin{figure}[h]
    \centering
    \includegraphics[width=1\linewidth]{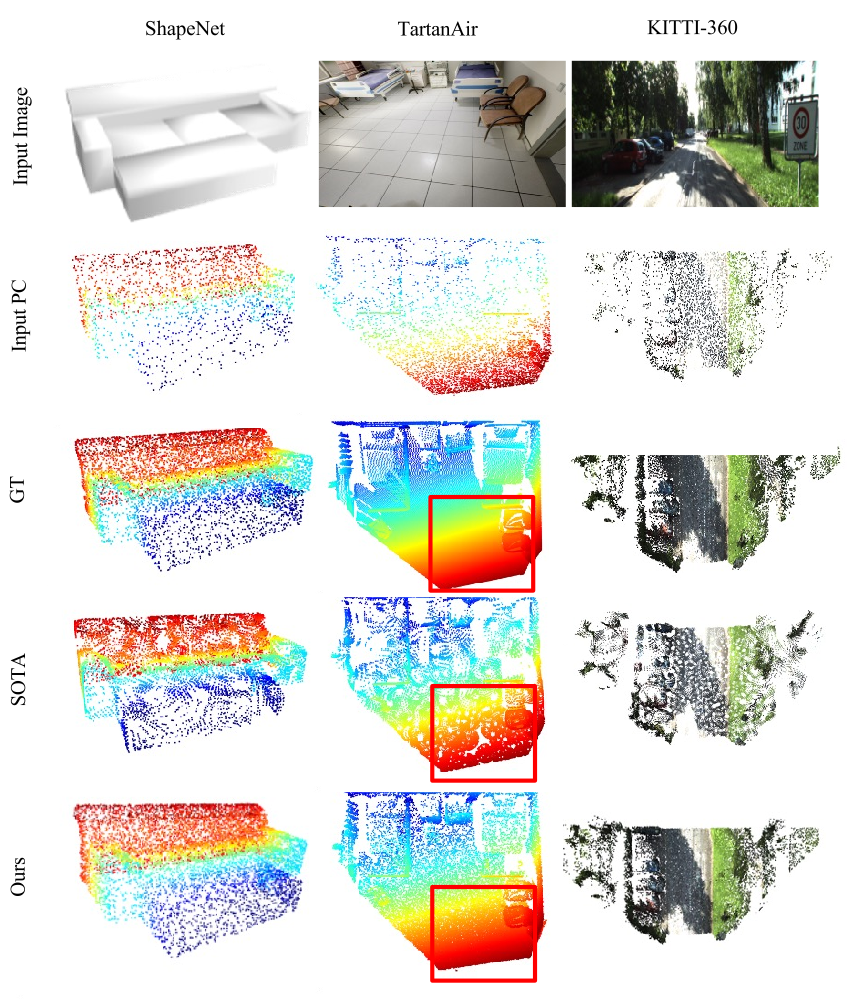} 
    \caption{Zoom-in figure of the upsampling task qualitative results.}
    \vspace{-46pt}
    \label{fig:upsampling_appendix}
\end{figure}

\begin{figure}[h]
    \centering
    \includegraphics[width=1\linewidth]{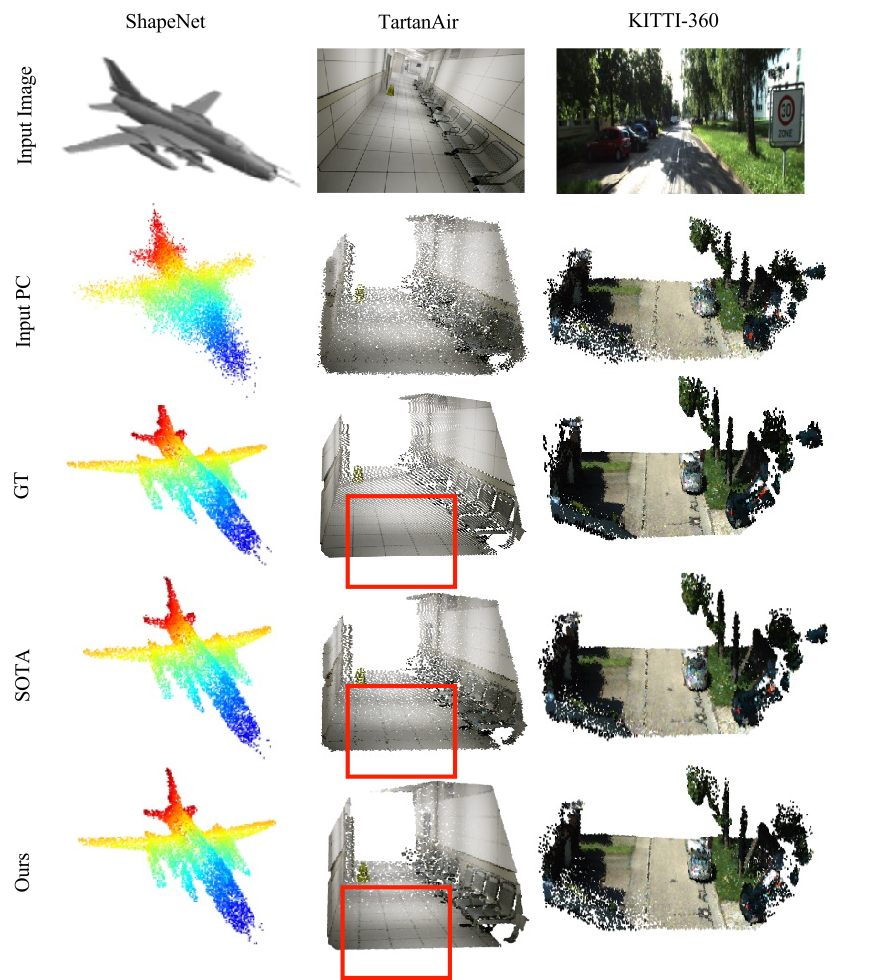} 
    \caption{Zoom-in figure of the denoising task qualitative results.}
    \vspace{-46pt}
    \label{fig:denoising_appendix}
\end{figure}

% \begin{figure*}[h]
%     \centering
%     \includegraphics[width=1\linewidth]{images/appendix1.pdf} 
%     \caption{More qualitative results on the ShapeNet Dataset.}
%     \vspace{16pt}
%     \label{fig:appendix1}
% \end{figure*}

% \begin{figure*}[h]
%     \centering
%     \includegraphics[width=1\linewidth]{images/appendix2.pdf} 
%     \caption{More qualitative results on the TartanAir Dataset.}
%     \vspace{16pt}
%     \label{fig:appendix2}
% \end{figure*}

% \begin{figure*}[t]
%     \centering
%     \includegraphics[width=1\linewidth]{images/appendix3.pdf} 
%     \caption{More qualitative results on the KITTI360 Dataset.}
%     \vspace{16pt}
%     \label{fig:appendix3}
% \end{figure*}

% % ---- Bibliography ----
% %
% % BibTeX users should specify bibliography style 'splncs04'.
% % References will then be sorted and formatted in the correct style.
% %
% \newpage
% \bibliography{main}

% \section{Rationale}
% \label{sec:rationale}
% % 
% Having the supplementary compiled together with the main paper means that:
% % 
% \begin{itemize}
% \item The supplementary can back-reference sections of the main paper, for example, we can refer to \cref{sec:intro};
% \item The main paper can forward reference sub-sections within the supplementary explicitly (e.g. referring to a particular experiment); 
% \item When submitted to arXiv, the supplementary will already included at the end of the paper.
% \end{itemize}
% % 
% To split the supplementary pages from the main paper, you can use \href{https://support.apple.com/en-ca/guide/preview/prvw11793/mac#:~:text=Delete%20a%20page%20from%20a,or%20choose%20Edit%20%3E%20Delete).}{Preview (on macOS)}, \href{https://www.adobe.com/acrobat/how-to/delete-pages-from-pdf.html#:~:text=Choose%20%E2%80%9CTools%E2%80%9D%20%3E%20%E2%80%9COrganize,or%20pages%20from%20the%20file.}{Adobe Acrobat} (on all OSs), as well as \href{https://superuser.com/questions/517986/is-it-possible-to-delete-some-pages-of-a-pdf-document}{command line tools}.

\end{document}